\documentclass[11pt]{article}
\usepackage{coling2020}
\usepackage{times}
\usepackage{url}
\usepackage{latexsym}

\colingfinalcopy 

\usepackage{amsmath,amsfonts,bm}

\def\eqref#1{equation~\ref{#1}}

\def\1{\bm{1}}

\DeclareMathAlphabet{\mathsfit}{\encodingdefault}{\sfdefault}{m}{sl}
\SetMathAlphabet{\mathsfit}{bold}{\encodingdefault}{\sfdefault}{bx}{n}

\usepackage{makecell}
\usepackage{xspace}
\usepackage{amsmath}
\usepackage{multirow}
\usepackage{booktabs}
\usepackage{graphicx}
\usepackage{wrapfig}
\usepackage{algorithm,algorithmic}
\usepackage{eqparbox}

\title{Learning to Few-Shot Learn \\Across Diverse Natural Language Classification Tasks}

\newcommand*\samethanks[1][\value{footnote}]{\footnotemark[#1]}

\author{
Trapit Bansal\thanks{$\;$Equal Contribution. Correspondence: \texttt{tbansal@cs.umass.edu}}$\;^\dagger$ \and Rishikesh Jha\samethanks$\;^\ddagger$ \and Andrew McCallum$^\dagger$ \\
$^\dagger$University of Massachusetts, Amherst \\
$^\ddagger$Code for Science and Society \\
\texttt{\{tbansal,rishikeshjha,mccallum\}@cs.umass.edu}
}
\newcommand{\method}{LEOPARD\xspace}

\begin{document}

\maketitle

\begin{abstract}
Pre-trained transformer models have shown enormous success in improving performance on several downstream tasks.
However, fine-tuning on a new task still requires large amounts of task-specific labeled data to achieve good performance.
We consider this problem of learning to generalize to new tasks with a few examples as a meta-learning problem.
While meta-learning has shown tremendous progress in recent years, its application is still limited to simulated problems or problems with limited diversity across tasks.
We develop a novel method, \method, which enables optimization-based meta-learning across tasks with different number of classes, 
and evaluate different methods on generalization to diverse NLP classification tasks.
\method is trained with the state-of-the-art transformer architecture and shows better generalization to tasks not seen at all during training, with as few as 4 examples per label.
Across 17 NLP tasks, including diverse domains of entity typing, natural language inference, sentiment analysis, and several other text classification tasks, we show that \method learns better initial parameters for few-shot learning than self-supervised pre-training or multi-task training, outperforming many strong baselines, for example, yielding 14.6\% average relative gain in accuracy on unseen tasks with only 4 examples per label.

\end{abstract}

\section{Introduction}
\blfootnote{
    %
    %
    \hspace{-0.65cm}  
    %
    %
    \hspace{-0.65cm}  
    This work is licensed under a Creative Commons 
    Attribution 4.0 International License.
    License details:
    \url{http://creativecommons.org/licenses/by/4.0/}.
}


Learning to learn \cite{schmidhuber1987evolutionary,bengio1992optimization,thrun2012learning} from limited supervision is an important problem with widespread application in areas where obtaining labeled data for training large models can be difficult or expensive. 
We consider this problem of \textit{learning in $k$-shots} for natural language processing (NLP) tasks, that is, given $k$ labeled examples of a new NLP task learn to efficiently solve the new task.
Recently, self-supervised pre-training of transformer models using language modeling objectives \cite{devlin2018bert,radford2019language,yang2019xlnet} has achieved tremendous success in learning general-purpose parameters which are useful for a variety of downstream NLP tasks.
While pre-training is beneficial, it is not optimized for fine-tuning with limited supervision and
such models still require large amounts of task-specific data for fine-tuning, in order to achieve good performance \cite{yogatama2018learning}.

On the other hand, meta-learning
methods have been proposed as effective solutions for few-shot learning.
Existing applications of such meta-learning methods have shown improved performance in few-shot learning for vision tasks such as learning to classify new image classes within a similar dataset. 
However, these applications are often limited to simulated datasets where each classification label is considered a task.
Moreover, their application in NLP has followed a similar trend \cite{han2018fewrel,yu2018diverse,guo2018multi,mi2019meta,geng2019induction}. 
Since the input space of natural language is shared across all NLP tasks, it is possible that a meta-learning approach generalizes to unseen tasks.
We thus move beyond simulated tasks to investigate meta-learning performance on generalization outside the training tasks, and focus on a diverse task-set with different number of labels across tasks.

Model agnostic meta-learning (MAML) \cite{Finn:2017:MMF:3305381.3305498} is an optimization-based approach to meta-learning which is agnostic to the model architecture and task specification. Hence, it is an ideal candidate for learning to learn from diverse tasks. However, it requires sharing model parameters, including softmax classification layers across tasks and learns a single initialization point across tasks.
This poses a barrier for learning across diverse tasks, 
where different tasks can have potentially disjoint label spaces. 
Contrary to this, multi-task learning \cite{caruana1997multitask} naturally handles disjoint label sets, while still benefiting from sharing statistical strength across tasks. 
However, to solve a new task, multi-task learning would require training a new classification layer for the task. 
On the other hand, metric-based approaches, such as prototypical networks \cite{vinyals2016matching,snell2017prototypical}, being non-parametric in nature can handle varied number of classes.
However, as the number of labeled examples increase, these methods do not adapt to leverage larger data and their performance can lag behind optimization-based methods.

We address these concerns and make the following contributions:
(1) we introduce a MAML-based meta-learning method, \textbf{\method}\footnote{\small\textbf{L}earning to g\textbf{e}nerate s\textbf{o}ftmax \textbf{pa}rameters fo\textbf{r} \textbf{d}iverse classification}, which is coupled with a parameter generator that learns to generate \textit{task-dependent} initial softmax classification parameters for any given task
and enables meta-learning across tasks with disjoint label spaces;
(2) we train \method with a transformer model, BERT \cite{devlin2018bert}, as the underlying neural architecture, and show that it is possible to learn better initialization parameters for few-shot learning than that obtained from just self-supervised pre-training or pre-training followed by multi-task learning;
(3) we evaluate on generalization, with a few-examples, to NLP tasks not seen during training or to new domains of seen tasks, including \textit{entity typing, natural language inference, sentiment classification, and various other text classification tasks};
(4) we study how meta-learning, multi-task learning and fine-tuning 
perform for few-shot learning of completely new tasks, analyze merits/demerits of parameter efficient meta-training, and study how various train tasks affect performance on target tasks.
To the best of our knowledge, this is the first application of meta-learning in NLP 
which evaluates on test tasks which are significantly different than training tasks and
goes beyond simulated classification tasks or domain-adaptation tasks (where train and test tasks are similar but from different domains).

\section{Background}
\label{sec:background}

In meta-learning, we consider a meta goal of learning across multiple tasks and assume a distribution over tasks $T_i \sim P(\mathcal{T})$. 
We follow the episodic learning framework of \newcite{vinyals2016matching} which minimizes train-test mismatch for few-shot learning.
We are given a set of $M$ training tasks $\{T_1, \ldots, T_M\}$, where each task instance potentially has a large amount of training data.
In order to simulate $k$-shot learning during training, in each episode (i.e.~a training step) a task $T_i$ is sampled with a training set $\mathcal{D}_i^{tr} \sim T_i$, consisting of only $k$ examples (per label) of the task and a validation set $\mathcal{D}_i^{val} \sim T_i$, containing several other examples of the same task.
The model $f$ is trained on $\mathcal{D}_i^{tr}$ using the task loss $\mathcal{L}_{i}$, and then evaluated on $\mathcal{D}_i^{val}$. The loss on $\mathcal{D}_i^{val}$ is then used to adjust the model parameters.
Here the validation error of the tasks serves as the training error for the meta-learning process.
At the end of training, the model is evaluated on a new task $T_{M+1} \sim P(\mathcal{T})$, where again the train set of $T_{M+1}$ contains only $k$ examples per label, and the model can use its learning procedure to adapt to the task $T_{M+1}$ using the train set.
We next discuss model-agnostic meta-learning which is pertinent to our work.


\textbf{Model-Agnostic Meta-Learning (MAML)}
\cite{Finn:2017:MMF:3305381.3305498} is an approach to optimization-based meta-learning where the goal is to find a good initial point for model parameters $\theta$, which through few steps of gradient descent, can be adapted to yield good performance on a new task.
Learning in MAML consists of an \textit{inner loop}, 
which applies gradient-based learning on the task-specific objective, and an \textit{outer-loop} which refines the initial point across tasks in order to enable fast learning.
Given a task $T_i$ with training datasets $\mathcal{D}_i^{tr}$ sampled during an episode, MAML's inner loop adapts the parameters $\theta$ as:
\begin{equation}
    \theta_i' = \theta - \alpha \nabla_{\theta}\mathcal{L}_{i}(\theta, \mathcal{D}_i^{tr}) \label{eq:inner_loop}
\end{equation}
Typically, more than one step of gradient update are applied sequentially. The learning-rate $\alpha$ can also be meta-learned in the outer-loop \cite{li2017meta}.
The parameters $\theta$ are then trained by back-propagating through the inner-loop adaptation, with the meta-objective of minimizing the error across respective task validation sets $\mathcal{D}_i^{val}$:
\begin{equation}
    \theta \leftarrow \theta - \beta \;\nabla_{\theta} \sum_{T_i \sim P(\mathcal{T})} \mathcal{L}_{i}(\theta_i',\mathcal{D}_i^{val}) \label{eq:outer_loop}
\end{equation}
Note that even though MAML is trained to generate a good initialization point for few-shot adaptation, since the inner-loop employs gradient-based learning, its performance can approach supervised learning in the limit of large data.

\section{Model}
\begin{wrapfigure}{r}{0.5\textwidth}
    \begin{center}
        \includegraphics[width=\linewidth]{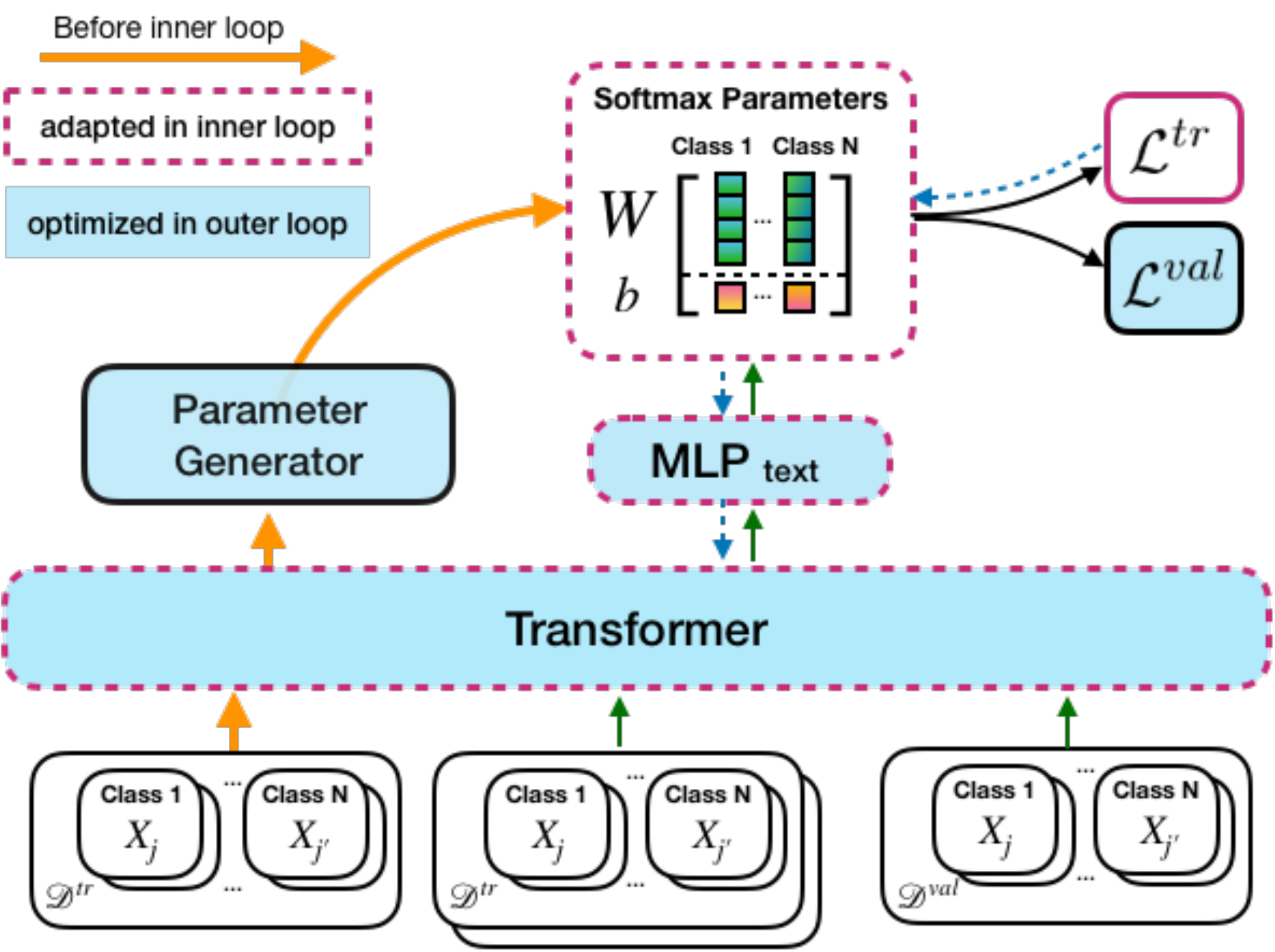}
    \end{center}
    \caption{The proposed \method model. Input is first encoded using the Transformer. The first batch from the support set is passed through the parameter generator which learns a per-class set representation that is used to generate the initial softmax parameters. Subsequently, the support batches are used for adaptation of the generated parameters as well as the encoder parameters. Pink box (dashed) outline shows modules that are adapted in the inner loop, whereas blue boxes are optimized in the outer loop.}
    \label{fig:leopard}
\end{wrapfigure}


\label{sec:method}

In this section, we describe our proposed method, \method, for learning new NLP classification tasks with $k$-examples.
Fig.~\ref{fig:leopard} shows a high-level description of the model.
Our approach builds on the MAML framework and addresses some of its limitations when applied to a diverse set of tasks with different number of classes across tasks.
Our model consists of three main components: 
(1) a shared neural input encoder which generates feature representations useful across tasks;
(2) a softmax parameter generator \textit{conditioned on the training dataset} for an $N$-way task, which generates the initial softmax parameters for the task;
(3) a MAML-based adaptation method with a distinction between \textit{task-specific parameters}, which are adapted per task, and \textit{task-agnostic} parameters, which are shared across tasks, that can lead to parameter-efficient fine-tuning of large models.
Full training algorithm is shown in Alg.~\ref{alg}.

\subsection{Text Encoder}
The input consists of natural language sentences, thus our models take sequences of words as input.
Note that some tasks require classifying pairs of sentences (such as natural language inference) and phrases in a sentence (such as entity typing), and we discuss how these can also be encoded as a sequence in Section \ref{sec:train_tasks}.
We use a Transformer model \cite{vaswani2017attention} as our text encoder which has shown success for many NLP tasks.
Concretely, we follow \newcite{devlin2018bert} and use their BERT-base model architecture. 
We denote the Transformer model by $f_{\theta}$, with parameters $\theta = \{\theta_1, \ldots, \theta_{12}\}$ where $\theta_v$ are the parameters of layer $v$. 
Transformer takes a sequence of words $\textbf{x}_j = [x_{j1}, \ldots, x_{jt}]$ as input ($t$ being the sequence length), and outputs $d$-dimensional contextualized representations at the final layer of multi-head self-attention.
BERT adds a special CLS token \cite{devlin2018bert} to the start of every input, which can be used as a sentence representation. We thus use this as the fixed-dimensional input feature representation of the sentence: $\tilde{x}_{j} = f_{\theta}([x_{j1}, \ldots, x_{js}])$.

\begin{algorithm}
 \caption{LEOPARD}
 \begin{algorithmic}[1]

  \REQUIRE set of $M$ training tasks and losses $\{(T_1, L_1), \ldots , (T_M, L_M)\}$, model parameters $\Theta = \{\theta, \psi, \alpha\}$, hyper-parameters $\nu, G, \beta$ \; \\
  Initialize $\theta$ with pre-trained BERT-base; \\
 \WHILE{not converged} 
 \STATE {\it \# sample batch of tasks}
 \FORALL{$T_i \in T$}
 \STATE $\mathcal{D}_i^{tr} \sim T_i$ \quad\quad {\it \# sample a batch of train data}
 \STATE $C_i^n \leftarrow \{x_j|y_j=n\}$ \quad\quad {\it \# partition data according to class labels}
 \STATE $w^n_i, b^n_i \leftarrow \frac{1}{|C_i^n|} \sum_{x_j \in C_i^n} g_{\psi}(f_{\theta}(x_j))$ \quad\quad {\it \# generate softmax parameters}
 \STATE $\textbf{W}_{i} \leftarrow [w^1_i; \ldots; w^{N_i}_i] $; \;\;$\textbf{b}_{i} \leftarrow [b^1_i; \ldots; b^{N_i}_i]$
 \STATE $\Phi_i^{(0)} \leftarrow \theta_{> \nu}\cup \{\phi, \textbf{W}_i, \textbf{b}_i\}$  \quad\quad {\it \# task-specific parameters}
 \FOR{$s := 0 \ldots G-1$} 
 \STATE  $\mathcal{D}_i^{tr} \sim T_i$ \quad\quad {\it \# sample a batch of train data}
\STATE $\Phi_i^{(s+1)} \leftarrow \Phi_i^{(s)} - \alpha_s\ \nabla_{\Phi}\mathcal{L}_{i}(\{\Theta, \Phi_i\}, \mathcal{D}_i^{tr})$ \quad\quad {\it \# adapt task-specific parameters}
 \ENDFOR
 \STATE $\mathcal{D}_i^{val} \sim T_i$ \quad\quad {\it \# sample a batch of validation data}
 \STATE $g_i \leftarrow  \nabla_{\Theta}\mathcal{L}_{i}(\{\Theta, \Phi_i^{(G)}\}, \mathcal{D}_i^{val})$ \quad\quad {\it \# gradient of task-agnostic parameters on validation}
 \ENDFOR
 \STATE $\Theta \leftarrow \Theta - \beta \cdot \sum_i g_i$ \quad\quad {\it \# optimize task-agnostic parameters}
 \ENDWHILE
\end{algorithmic}
\label{alg}
\end{algorithm}

\subsection{Generating Softmax Parameters for Task-specific Classification}
Existing applications of MAML consider few-shot learning with a fixed $N$, i.e.~ the number of classes.
This limits applicability to multiple types of tasks, each of which would require a different number of classes for classification. 
To remedy this, we introduce a method to generate \textit{task-dependent} softmax parameters (both linear weights and bias).
Given the training data, $\mathcal{D}_i^{tr} = \{(x_j, y_j)\}$, for a task $T_i$ in an episode, we first partition the input into the $N_i$ number of classes for the task (available in  $\mathcal{D}_i^{tr}$): $C_i^n = \{x_j | y_j = n\}$, where $n \in [N_i]$. 
Now, we perform a non-linear projection on the representations of the $x_j$ in each class partition obtained from the text-encoder, and obtain a set representation for class $n$:
\begin{equation}
    w^n_i, b^n_i = \frac{1}{|C_i^n|} \sum_{x_j \in C_i^n} g_{\psi}(f_{\theta}(\textbf{x}_j)) \label{eq:softmax}
\end{equation}
where $g_{\psi}$ is multi-layer perceptron (MLP) with two layers and $tanh$ non-linearities, $w_i^n$ is a $l$-dimensional vector and $b_i^n$ is a scalar. 
$w_i^n$ and $b_i^n$ are the softmax linear weight and bias, respectively, for the class $n$:
\begin{equation}
    \textbf{W}_{i} = [w^1_i; \ldots; w^{N_i}_i] \quad \textbf{b}_{i} = [b^1_i; \ldots; b^{N_i}_i]  \label{eq:softmax_final}
\end{equation}
Thus, the softmax classification weights $\textbf{W}_{i} \in \mathcal{R}^{N_i \times l}$ and bias $\textbf{b}_{i} \in \mathcal{R}^{N_i}$ for task $T_i$ are obtained by row-wise concatenation of the per-class weights in \eqref{eq:softmax}.
Note that encoder $g_{\psi}(\cdot)$ would be shared across tasks in different episodes.

Now, given the softmax parameters, the prediction for a new data-point $\textbf{x}^*$ is given as:
\begin{equation}
    p(y | \textbf{x}^*) = \mbox{softmax} \left\{ \textbf{W}_{i} h_\phi(f_{\theta}(\textbf{x}^*)) + \textbf{b}_{i} \right\} \label{eq:prediction}
\end{equation}
where $h_{\phi}(\cdot)$ is another MLP with parameters $\phi$ and output dimension $l$, and the softmax is over the set of classes $N_i$ for the task.

Note that if we use $x^* \in \mathcal{D}_i^{val}$, then the model is a form of a prototypical network \cite{snell2017prototypical} which uses a learned distance function.
However, this would limit the model to not adapt its parameters with increasing data.
We next discuss how we learn to adapt using the generated softmax.
It is important to note that \textit{we do not introduce any task-specific parameters}, unlike multi-task learning \cite{caruana1997multitask} which will require new softmax layers for each task, and the existing parameters are used to generate a good starting point for softmax parameters across tasks which can then be \textit{adapted} 
using stochastic gradient (SGD) based learning.

\subsection{Learning to Adapt Efficiently}
\label{sec:adapt}
Given the task-specific classification loss computed at an episode, MAML takes multiple steps of SGD on the same training set $\mathcal{D}_i^{tr}$, as in \eqref{eq:inner_loop}.
We apply MAML on the model parameters, including the generated softmax parameters.
However, the number of parameters in BERT 
is substantially high ($\sim110$ million) and it can be beneficial to adapt a smaller number of parameters \cite{houlsby2019parameter,zintgraf2018cavia}.
We thus separate the set of parameters into \textit{task-specific} and \textit{task-agnostic}.
For the transformer parameters for each layer $\{\theta_v\}$, we consider a threshold $\nu$ over layers, and consider $\theta_{\leq \nu} = \{\theta_v | v \leq \nu\}$ to be the parameters for first $\nu$ layers (closest to the input) and the rest of the parameters as $\theta_{> \nu}$.
Then we consider $\theta_{\leq \nu}$
and the parameters $\psi$ of the softmax generating function (\eqref{eq:softmax}) as the set of task-agnostic parameters $\Theta = \theta_{\le \nu}\cup \{\psi\}$. 
These task-agnostic parameters $\Theta$ need to generalize to produce good feature representations and good initial point for classification layer \textit{across tasks}.
The remaining set of parameters for the higher layers of transformer, the input projection function in \ref{eq:prediction}, and the softmax weights and bias \textit{generated} in \eqref{eq:softmax_final}
are considered as the set of task-specific parameters $\Phi_i = \theta_{> \nu}\cup \{\phi, \textbf{W}_i, \textbf{b}_i\}$.

The task-specific parameters will be adapted for each task using SGD, as in \eqref{eq:inner_loop}.
Note that MAML usually does gradient descent steps on the same meta-train batch $\mathcal{D}_i^{tr}$ for a task in an episode.
However, since we use $\mathcal{D}_i^{tr}$ to generate the softmax parameters in \eqref{eq:softmax}, using the same data to also take multiple gradient steps can lead to over-fitting.
Thus, we instead sample $G > 1$ meta-train batches in each episode of training,
and use the subesequent batches (after the first batch) for adaptation.
Task-specific adaptation in the inner loop does $G$ steps of the following update, starting with $\Phi_i^{(0)} \leftarrow \Phi_i$, for $s := 0, \ldots, G -1$:
\begin{equation}
    \Phi_i^{(s+1)} = \Phi_i^{(s)} - \alpha_s\; \mathbb{E}_{\mathcal{D}_i^{tr} \sim T_i}\left[\nabla_{\Phi}\mathcal{L}_{i}(\{\Theta, \Phi_i\}, \mathcal{D}_i^{tr})\right] 
    \label{inner_loop_sgd}
\end{equation}
Note that we only take gradient with respect to the task-specific parameters $\Phi_i$, however the updated parameter is also a function of $\Theta$.
After the $G$ steps of adaptation, the final point (which consists of parameters $\Theta$ and $\Phi^G$) is evaluated on the validation set for the task, $\mathcal{D}_i^{val}$, and the task-agnostic parameters $\Theta$ are updated (as in \eqref{eq:outer_loop}) to adjust the initial point across tasks. 
Note that optimization of the task-agnostic parameters requires back-propagating through the inner-loop gradient steps and requires computing higher-order gradients.
\newcite{Finn:2017:MMF:3305381.3305498} proposed using a first-order approximation for computational efficiency. 
We use this approximation in this work, however we note that the distinction between task-specific and task-agnostic parameters can allow for higher order gradients when there are few task-specific parameters (for example, only the last layer). 

\textbf{Other Technical Details: } 
For few-shot learning, learning rate can often be an important hyper-parameter and the above approach can benefit from also learning the learning-rate for adaptation \cite{li2017meta}.
Instead of scalar inner loop learning rates, it has been shown beneficial to have per-parameter learning rates that are also learned \cite{li2017meta,antoniou2018train}.
However, this doubles the number of parameters and can be inefficient.
Instead, we learn a per-layer learning rate for the inner loop to allow different transformer layers to adapt at different rates.
We apply layer normalization across layers of transformers \cite{vaswani2017attention,ba2016layer} and also adapt their parameters in the inner loop.
The number of layers to consider as task-specific, $\nu$, is a hyper-parameter.
We initialize the meta-training of \method from pre-trained BERT model which stabilizes training.



\section{Experiments}
\label{sec:exp}
Our experiments evaluate how different methods generalize to new NLP tasks with limited supervision.
We focus on sentence-level classification tasks, including natural language inference (NLI) tasks which require classifying pairs of sentences as well as tasks like entity typing which require classifying a phrase in a sentence.
We consider 17 target tasks\footnote{Code, trained model parameters, and datasets: \url{https://github.com/iesl/leopard}}. 
Main results are in Sec.~\ref{sec:results}.

\subsection{Training Tasks} 
\label{sec:train_tasks}
We use the GLUE benchmark tasks
\cite{wang2018glue} for training all the models.
Such tasks are considered important for general linguistic intelligence, have lots of supervised data for many tasks and have been useful for transfer learning \cite{phang2018sentence,wang2018can}.
We consider the following tasks for training\footnote{We exclude WNLI since its training data is small and STS-B task since it is a regression task}: MNLI (m/mm), SST-2, QNLI, QQP, MRPC, RTE, and the SNLI dataset \cite{bowman2015snli}.
We use the corresponding validation sets for hyper-parameter tuning and early stopping.
For meta-learning methods, we classify between every pair of labels (for tasks with more than 2 labels) which increases the number of tasks and allows for more per-label examples in a batch during training. 
Moreover, to learn to do phrase-level classification, we modify SST (for all models) which is a phrase-level sentiment classification task by providing a sentence in which the phrase occurs as part of the input. That is, the input is the sentence followed by a separator token \cite{devlin2018bert} followed by the phrase to classify. 
See Appendix \ref{sec:moredata} for more details.

\subsection{Evaluation and Baselines}
Unlike existing methods which evaluate meta-learning models on sampled tasks from a fixed dataset \cite{vinyals2016matching,Finn:2017:MMF:3305381.3305498}, we evaluate methods on real NLP datasets by using the entire test sets for the target task after using a sampled $k$-shot training data for fine-tuning.
The models parameters are trained on the set of training tasks and are then fine-tuned with $k$ training examples per label for a target test task.
The fine-tuned models are then evaluated on the \textit{entire test-set} for the task.
We evaluate on $k \in \{4, 8, 16\}$.
For each task, for every $k$, we sample 10 training datasets and report the mean and standard deviation, since model performance can be sensitive to the $k$ examples chosen for training.
In the few-shot setting it can be unreasonable to assume access to a large validation set \cite{yu2018diverse,kann2019towards}, thus for the fine-tuning step we tuned the hyper-parameters for all baselines on a held out validation task. 
We used SciTail, a scientific NLI task, and electronics domain of Amazon sentiment classification task as the validation tasks. 
We took the hyper-parameters that gave best average performance on validation data of these tasks, for each value of $k$.
For \method, we only tune the number of epochs for fine-tuning, use the learned per-layer learning rates and reuse remaining hyper-parameters (see Appendix \ref{sec:hyper}).

We evaluate multiple transfer learning baselines as well as a meta-learning baseline.
Note that most existing applications of few-shot learning are tailored towards specific tasks and don't trivially apply to diverse tasks considered here. We evaluate the following methods:\\
\textbf{$\mbox{BERT}_{\mbox{base}}$}: We use the cased BERT-base model \cite{devlin2018bert} which is a state-of-the-art transformer \cite{vaswani2017attention} model for NLP. BERT uses language model pre-training followed by supervised fine-tuning on a downstream task. 
For fine-tuning, we tune all parameters
as it performed better on the validation task. \\
\textbf{Multi-task BERT ($\mbox{MT-BERT}$)}: This is the BERT-base model trained in a multi-task learning setting on the set of training tasks. Our MT-BERT is comparable to the MT-DNN model of \newcite{liu2019multi} that is trained on the tasks considered here and uses the cased BERT-base as the initialization. We did not use the specialized stochastic answer network for NLI used by MT-DNN. 
For this model, we tune all the parameters during fine-tuning. \\
\textbf{$\mbox{MT-BERT}_{\mbox{softmax}}$}: This is the multi-task BERT model above, where we only tune the softmax layer during fine-tuning. \\
\textbf{Prototypical BERT (Proto-BERT)}: This is the prototypical network method \cite{snell2017prototypical} that uses BERT-base as the underlying neural model. 
Following \newcite{snell2017prototypical}, we used euclidean distance as the distance metric.\\
All methods are initialized with pre-trained BERT. All parameters of MT-BERT and Proto-BERT are also tuned during training.
We don't compare with MAML \cite{Finn:2017:MMF:3305381.3305498} as it does not trivially support varying number of classes, and show in ablations (\ref{sec:ablation}) that solutions like using zero-initialized initial softmax perform worse.

\noindent \textbf{Implementation Details}:
Since dataset sizes can be imbalanced, it can affect multi-task and meta-learning performance.
\newcite{wang2018can} analyze this in detail for multi-task learning. We explored sampling tasks with uniform probability, proportional to size and proportional to the square-root of the size of the task. 
For all models, we found the latter to be beneficial.
All methods are trained on 4 GPUs to benefit from large batches.
Best hyper-parameters, search ranges and data statistics are in Appendix. 

\subsection{Results}
\label{sec:results}
We evaluate all the models on 17 target NLP tasks.
None of the task data is observed during the training of the models, and the models are fine-tuned on few examples for the target task and then evaluated on the entire test set for the task.
For $k$-shot learning of tasks not seen at all during training, we observe, on average, relative gain in accuracy of $14.60\%$, $10.83\%$, and $11.16\%$,
for $k=4,8,16$ respectively.

\subsubsection{Generalization Beyond Training Tasks}

\begin{table*}[htb!]
\centering \fontsize{8.0}{9.5}\selectfont \setlength{\tabcolsep}{0.5em}
\begin{tabular}{cccccccc}
\Xhline{2\arrayrulewidth}
\multicolumn{8}{c}{\textbf{Entity Typing}}    \\ \Xhline{2\arrayrulewidth}
         & $N$ & $k$ &$\mbox{BERT}_{\mbox{base}}$ & $\mbox{MT-BERT}_{\mbox{softmax}}$ & MT-BERT &  Proto-BERT & \method \\[3pt] \Xhline{2\arrayrulewidth}
\multirow{3}{*}{CoNLL} & \multirow{3}{*}{4} & 4 & 50.44 \tiny{$\pm$ 08.57} & 52.28 \tiny{$\pm$ 4.06} & \textbf{55.63} \tiny{$\pm$ 4.99} & 32.23 \tiny{$\pm$ 5.10} & 54.16 \tiny{$\pm$ 6.32} \\
& & 8 & 50.06 \tiny{$\pm$ 11.30} &  65.34 \tiny{$\pm$ 7.12} & 58.32 \tiny{$\pm$ 3.77} & 34.49 \tiny{$\pm$ 5.15} & \textbf{67.38} \tiny{$\pm$ 4.33} \\
& & 16 & 74.47 \tiny{$\pm$ 03.10} & 71.67 \tiny{$\pm$ 3.03} & 71.29 \tiny{$\pm$ 3.30} & 33.75 \tiny{$\pm$ 6.05} & \textbf{76.37} \tiny{$\pm$ 3.08} \\
\midrule
\multirow{3}{*}{MITR} & \multirow{3}{*}{8} & 4 & 49.37 \tiny{$\pm$ 4.28} & 45.52 \tiny{$\pm$ 5.90} & \textbf{50.49} \tiny{$\pm$ 4.40} & 17.36 \tiny{$\pm$ 2.75} & 49.84 \tiny{$\pm$ 3.31} \\
& & 8 & 49.38 \tiny{$\pm$ 7.76} & 58.19 \tiny{$\pm$ 2.65} & 58.01 \tiny{$\pm$ 3.54} & 18.70 \tiny{$\pm$ 2.38} & \textbf{62.99} \tiny{$\pm$ 3.28} \\
& & 16 & 69.24 \tiny{$\pm$ 3.68} & 66.09 \tiny{$\pm$ 2.24} & 66.16 \tiny{$\pm$ 3.46} & 16.41 \tiny{$\pm$ 1.87} & \textbf{70.44} \tiny{$\pm$ 2.89} \\

\Xhline{2\arrayrulewidth}
\multicolumn{8}{c}{\textbf{Text Classification}}\\ \Xhline{2\arrayrulewidth}
\multirow{3}{*}{Airline} & \multirow{3}{*}{3} & 4 &42.76 \tiny{$\pm$ 13.50} & 43.73 \tiny{$\pm$ 7.86} & 46.29 \tiny{$\pm$ 12.26} & 40.27 \tiny{$\pm$ 8.19} & \textbf{54.95} \tiny{$\pm$ 11.81} \\
& & 8 &38.00 \tiny{$\pm$ 17.06} & 52.39 \tiny{$\pm$ 3.97} & 49.81 \tiny{$\pm$ 10.86} & 51.16 \tiny{$\pm$ 7.60} & \textbf{61.44} \tiny{$\pm$ 03.90} \\
& & 16 & 58.01 \tiny{$\pm$ 08.23} & 58.79 \tiny{$\pm$ 2.97} & 57.25 \tiny{$\pm$ 09.90} & 48.73 \tiny{$\pm$ 6.79} & \textbf{62.15} \tiny{$\pm$ 05.56} \\
\midrule
\multirow{3}{*}{Disaster} & \multirow{3}{*}{2} & 4 & \textbf{55.73} \tiny{$\pm$ 10.29} & 52.87 \tiny{$\pm$ 6.16} & 50.61 \tiny{$\pm$ 8.33} & 50.87 \tiny{$\pm$ 1.12} & 51.45 \tiny{$\pm$ 4.25} \\
 & & 8 & \textbf{56.31} \tiny{$\pm$ 09.57} & 56.08 \tiny{$\pm$ 7.48} & 54.93 \tiny{$\pm$ 7.88} & 51.30 \tiny{$\pm$ 2.30} & 55.96 \tiny{$\pm$ 3.58} \\
 & & 16 & 64.52 \tiny{$\pm$ 08.93} & \textbf{65.83} \tiny{$\pm$ 4.19} & 60.70 \tiny{$\pm$ 6.05} & 52.76 \tiny{$\pm$ 2.92} & 61.32 \tiny{$\pm$ 2.83} \\
\midrule
\multirow{3}{*}{Emotion} & \multirow{3}{*}{13} & 4 & 09.20 \tiny{$\pm$ 3.22} & 09.41 \tiny{$\pm$ 2.10} & 09.84 \tiny{$\pm$ 2.14} & 09.18 \tiny{$\pm$ 3.14} & \textbf{11.71} \tiny{$\pm$ 2.16} \\
 && 8 & 08.21 \tiny{$\pm$ 2.12} & 11.61 \tiny{$\pm$ 2.34} & 11.21 \tiny{$\pm$ 2.11} & 11.18 \tiny{$\pm$ 2.95} & \textbf{12.90} \tiny{$\pm$ 1.63} \\
 && 16 & 13.43 \tiny{$\pm$ 2.51} & \textbf{13.82} \tiny{$\pm$ 2.02} & 12.75 \tiny{$\pm$ 2.04} & 12.32 \tiny{$\pm$ 3.73} & 13.38 \tiny{$\pm$ 2.20} \\
\midrule
\multirow{3}{*}{Political Bias} & \multirow{3}{*}{2} & 4 &54.57 \tiny{$\pm$ 5.02} & 54.32 \tiny{$\pm$ 3.90} & 54.66 \tiny{$\pm$ 3.74} & 56.33 \tiny{$\pm$ 4.37} & \textbf{60.49} \tiny{$\pm$ 6.66} \\
& & 8 &56.15 \tiny{$\pm$ 3.75} & 57.36 \tiny{$\pm$ 4.32} & 54.79 \tiny{$\pm$ 4.19} & 58.87 \tiny{$\pm$ 3.79} & \textbf{61.74} \tiny{$\pm$ 6.73} \\
& & 16 &60.96 \tiny{$\pm$ 4.25} & 59.24 \tiny{$\pm$ 4.25} & 60.30 \tiny{$\pm$ 3.26} & 57.01 \tiny{$\pm$ 4.44} & \textbf{65.08} \tiny{$\pm$ 2.14} \\
\midrule
\multirow{3}{*}{Political Audience} & \multirow{3}{*}{2} & 4 & 51.89 \tiny{$\pm$ 1.72} & 51.50 \tiny{$\pm$ 2.72} & 51.53 \tiny{$\pm$ 1.80} & 51.47 \tiny{$\pm$ 3.68} & \textbf{52.60} \tiny{$\pm$ 3.51} \\
 & & 8 &52.80 \tiny{$\pm$ 2.72} & 53.53 \tiny{$\pm$ 2.26} & \textbf{54.34} \tiny{$\pm$ 2.88} & 51.83 \tiny{$\pm$ 3.77} & 54.31 \tiny{$\pm$ 3.95} \\
 & & 16 & \textbf{58.45} \tiny{$\pm$ 4.98} & 56.37 \tiny{$\pm$ 2.19} & 55.14 \tiny{$\pm$ 4.57} & 53.53 \tiny{$\pm$ 3.25} & 57.71 \tiny{$\pm$ 3.52} \\
 \midrule
\multirow{3}{*}{Political Message} & \multirow{3}{*}{9} & 4 &15.64 \tiny{$\pm$ 2.73} & 13.71 \tiny{$\pm$ 1.10} & 14.49 \tiny{$\pm$ 1.75} & 14.22 \tiny{$\pm$ 1.25} & \textbf{15.69} \tiny{$\pm$ 1.57} \\
 & & 8 &13.38 \tiny{$\pm$ 1.74} & 14.33 \tiny{$\pm$ 1.32} & 15.24 \tiny{$\pm$ 2.81} & 15.67 \tiny{$\pm$ 1.96} & \textbf{18.02} \tiny{$\pm$ 2.32} \\
 & & 16 & \textbf{20.67} \tiny{$\pm$ 3.89} & 18.11 \tiny{$\pm$ 1.48} & 19.20 \tiny{$\pm$ 2.20} & 16.49 \tiny{$\pm$ 1.96} & 18.07 \tiny{$\pm$ 2.41} \\
\midrule

\multirow{3}{*}{Rating Books} & \multirow{3}{*}{3} & 4 & 39.42 \tiny{$\pm$ 07.22} & 44.82 \tiny{$\pm$ 9.00} & 38.97 \tiny{$\pm$ 13.27} & 48.44 \tiny{$\pm$ 7.43} & \textbf{54.92} \tiny{$\pm$ 6.18} \\
 & & 8 &39.55 \tiny{$\pm$ 10.01} & 51.14 \tiny{$\pm$ 6.78} & 46.77 \tiny{$\pm$ 14.12} & 52.13 \tiny{$\pm$ 4.79} & \textbf{59.16} \tiny{$\pm$ 4.13} \\
 & & 16 &43.08 \tiny{$\pm$ 11.78} & 54.61 \tiny{$\pm$ 6.79} & 51.68 \tiny{$\pm$ 11.27} & 57.28 \tiny{$\pm$ 4.57} & \textbf{61.02} \tiny{$\pm$ 4.19} \\
\midrule
\multirow{3}{*}{Rating DVD} & \multirow{3}{*}{3} & 4 &32.22 \tiny{$\pm$ 08.72} & 45.94 \tiny{$\pm$ 7.48} & 41.23 \tiny{$\pm$ 10.98} & 47.73 \tiny{$\pm$ 6.20} & \textbf{49.76} \tiny{$\pm$ 9.80} \\
 & & 8 &36.35 \tiny{$\pm$ 12.50} & 46.23 \tiny{$\pm$ 6.03} & 45.24 \tiny{$\pm$ 9.76} & 47.11 \tiny{$\pm$ 4.00} & \textbf{53.28} \tiny{$\pm$ 4.66} \\
 & & 16 &42.79 \tiny{$\pm$ 10.18} & 49.23 \tiny{$\pm$ 6.68} & 45.19 \tiny{$\pm$ 11.56} & 48.39 \tiny{$\pm$ 3.74} & \textbf{53.52} \tiny{$\pm$ 4.77} \\
\midrule
\multirow{3}{*}{Rating Electronics} & \multirow{3}{*}{3} & 4 &39.27 \tiny{$\pm$ 10.15} & 39.89 \tiny{$\pm$ 5.83} & 41.20 \tiny{$\pm$ 10.69} & 37.40 \tiny{$\pm$ 3.72} & \textbf{51.71} \tiny{$\pm$ 7.20} \\
 & & 8 &28.74 \tiny{$\pm$ 08.22} & 46.53 \tiny{$\pm$ 5.44} & 45.41 \tiny{$\pm$ 09.49} & 43.64 \tiny{$\pm$ 7.31} & \textbf{54.78} \tiny{$\pm$ 6.48} \\
 & & 16 &45.48 \tiny{$\pm$ 06.13} & 48.71 \tiny{$\pm$ 6.16} & 47.29 \tiny{$\pm$ 10.55} & 44.83 \tiny{$\pm$ 5.96} & \textbf{58.69} \tiny{$\pm$ 2.41} \\
\midrule
\multirow{3}{*}{Rating Kitchen} & \multirow{3}{*}{3} & 4 &34.76 \tiny{$\pm$ 11.20} & 40.41 \tiny{$\pm$ 5.33} & 36.77 \tiny{$\pm$ 10.62} & 44.72 \tiny{$\pm$ 9.13} & \textbf{50.21} \tiny{$\pm$ 09.63} \\
 & & 8 &34.49 \tiny{$\pm$ 08.72} & 48.35 \tiny{$\pm$ 7.87} & 47.98 \tiny{$\pm$ 09.73} & 46.03 \tiny{$\pm$ 8.57} & \textbf{53.72} \tiny{$\pm$ 10.31} \\
 & & 16 &47.94 \tiny{$\pm$ 08.28} & 52.94 \tiny{$\pm$ 7.14} & 53.79 \tiny{$\pm$ 09.47} & 49.85 \tiny{$\pm$ 9.31} & \textbf{57.00} \tiny{$\pm$ 08.69} \\[1pt]
\Xhline{3\arrayrulewidth}\rule{0mm}{3mm}
\multirow{3}{*}{Overall Average} &  & 4 & 38.13 & 40.13 & 40.10 & 36.29 & \textbf{45.99} \\
     & & 8 & 36.99 & 45.89 & 44.25 & 39.15 & \textbf{50.86} \\
 & & 16 & 48.55 & 49.93 & 49.07 & 39.85 & \textbf{55.50} \\
\bottomrule
\end{tabular}
\caption{Few-shot generalization performance across tasks not seen during training. $k$ is the number of examples per label for fine-tuning and $N$ is the number of classes for the task. On average, \method is significantly better than other models for few-shot transfer to new tasks.}
\label{tab:general}
\end{table*}

\begin{table*}[htb]
\centering \fontsize{8.0}{9.5}\selectfont \setlength{\tabcolsep}{0.5em}
\begin{tabular}{cccccccc}
\Xhline{2\arrayrulewidth}
\multicolumn{8}{c}{\textbf{Natural Language Inference}}                                              \\ \Xhline{2\arrayrulewidth}
 & $k$ & $\mbox{BERT}_{\mbox{base}}$ & $\mbox{MT-BERT}_{\mbox{softmax}}$ & MT-BERT & $\mbox{MT-BERT}_{\mbox{reuse}}$ & Proto-BERT & \method \\[3pt] \Xhline{2\arrayrulewidth}
\multirow{3}{*}{Scitail}    & 4 & 58.53 \tiny{$\pm$ 09.74} & 74.35 \tiny{$\pm$ 5.86} & 63.97 \tiny{$\pm$ 14.36} & \textbf{76.65 \tiny{$\pm$ 2.45}} & 76.27 \tiny{$\pm$ 4.26} & 69.50 \tiny{$\pm$ 9.56} \\
 & 8 &57.93 \tiny{$\pm$ 10.70} & \textbf{79.11 \tiny{$\pm$ 3.11}} & 68.24 \tiny{$\pm$ 10.33} & 76.86 \tiny{$\pm$ 2.09} & 78.27 \tiny{$\pm$ 0.98} & 75.00 \tiny{$\pm$ 2.42} \\
 & 16 &65.66 \tiny{$\pm$ 06.82} & \textbf{79.60 \tiny{$\pm$ 2.31}} & 75.35 \tiny{$\pm$ 04.80} & 79.53 \tiny{$\pm$ 2.17} & 78.59 \tiny{$\pm$ 0.48} & 77.03 \tiny{$\pm$ 1.82} \\
\Xhline{2\arrayrulewidth}
\multicolumn{8}{c}{\textbf{Amazon Review Sentiment Classification}}    \\ \Xhline{2\arrayrulewidth}
\multirow{3}{*}{Books} & 4 &54.81 \tiny{$\pm$ 3.75} & 68.69 \tiny{$\pm$ 5.21} & 64.93 \tiny{$\pm$ 8.65} & 74.79 \tiny{$\pm$ 6.91} & 73.15 \tiny{$\pm$ 5.85} & \textbf{82.54 \tiny{$\pm$ 1.33}} \\
 & 8 &53.54 \tiny{$\pm$ 5.17} & 74.86 \tiny{$\pm$ 2.17} & 67.38 \tiny{$\pm$ 9.78} & 78.21 \tiny{$\pm$ 3.49} & 75.46 \tiny{$\pm$ 6.87} & \textbf{83.03 \tiny{$\pm$ 1.28}} \\
 & 16 &65.56 \tiny{$\pm$ 4.12} & 74.88 \tiny{$\pm$ 4.34} & 69.65 \tiny{$\pm$ 8.94} & 78.87 \tiny{$\pm$ 3.32} & 77.26 \tiny{$\pm$ 3.27} & \textbf{83.33 \tiny{$\pm$ 0.79}} \\
                    \midrule
\multirow{3}{*}{Kitchen} & 4 &56.93 \tiny{$\pm$ 7.10} & 63.07 \tiny{$\pm$ 7.80} & 60.53 \tiny{$\pm$ 9.25} & 75.40 \tiny{$\pm$ 6.27} & 62.71 \tiny{$\pm$ 9.53} & \textbf{78.35 \tiny{$\pm$ 18.36}} \\
 & 8 &57.13 \tiny{$\pm$ 6.60} & 68.38 \tiny{$\pm$ 4.47} & 69.66 \tiny{$\pm$ 8.05} & 75.13 \tiny{$\pm$ 7.22} & 70.19 \tiny{$\pm$ 6.42} & \textbf{84.88 \tiny{$\pm$ 01.12}} \\
 & 16 &68.88 \tiny{$\pm$ 3.39} & 75.17 \tiny{$\pm$ 4.57} & 77.37 \tiny{$\pm$ 6.74} & 80.88 \tiny{$\pm$ 1.60} & 71.83 \tiny{$\pm$ 5.94} & \textbf{85.27 \tiny{$\pm$ 01.31}} \\
  \midrule
\bottomrule
\end{tabular}
\caption{Domain transfer evaluation (accuracy) on NLI and Sentiment classification datasets.}
\label{tab:nli}
\end{table*}
We use the following datasets (more details in Appendix): (1) entity typing: CoNLL-2003 \cite{sang2003conll}, MIT-Restaurant \cite{liu2013asgard}; 
(2) rating classification: we use the review ratings for each domain from the Amazon Reviews dataset \cite{blitzer2007biographies} and consider a 3-way classification based on the ratings;
(3) text classification: social-media datasets from crowdflower\footnote{https://www.figure-eight.com/data-for-everyone/}.

Table \ref{tab:general} shows the performance. 
We can see that, on average, \method outperforms all the baselines, yielding significant improvements in accuracy.
This shows \method's robustness to varying number of labels across tasks and across different text domains. 
Note that \method uses the same training tasks as $\mbox{MT-BERT}$ but can adapt to new tasks with fewer examples, and improvements are highest with only 4 examples.
Performance of prototypical networks is worse than most other fine-tuning methods on new training tasks. 
We hypothesize that this is because prototypical networks do not generate good class prototypes for new tasks and adaptation of class prototypes is important for improving performance. We also see that improved feature learning in $\mbox{MT-BERT}$ with additional training tasks serves as a better initialization point for held-out tasks than BERT, and only tuning the softmax layer of this model is slightly better than tuning all parameters.
Interestingly, on some tasks like Disaster classification, we observe BERT to perform better than other models, indicating negative transfer from the training tasks.

\subsubsection{Few-Shot Domain Transfer}
We now evaluate performance on new domains of tasks seen at training time.
For this, we consider two tasks of Sentiment Classification and NLI.
For sentiment classification we use 4 domains of Amazon reviews \cite{blitzer2007biographies} 
and for NLI we use a scientific entailment dataset (SciTail) \cite{khot2018scitail}.
We introduce another relevant baseline here, MT-BERT$_{\mbox{reuse}}$, which reuses the trained softmax parameters of a related train task.
Results
are summarized in Table \ref{tab:nli}, we show two domains of sentiment classification and more results are in Appendix \ref{sec:more_results}.
Note that the related train task, SST, only contains phrase-level sentiments and the models weren't trained to predict sentence-level sentiment, while the target tasks require sentence-level sentiment. 
We observe that \method performs better than the baselines on all domains of sentiment classification, while on Scitail MT-BERT models perform better, potentially because training consisted of many related NLI datasets.
Note that prototypical networks is a competitive baseline here and its performance is better for these tasks in comparison to those in Table \ref{tab:general} as it has learned to generate prototypes for a similar task during training.

\subsection{Ablation Study}
\label{sec:ablation}
\begin{table*}[tb!]
\centering 
\fontsize{8.8}{10.5}\selectfont \setlength{\tabcolsep}{0.5em}
\begin{tabular}{clccc}
\Xhline{2\arrayrulewidth}
     $k$    &   \multicolumn{1}{c}{Model}   & Entity Typing & Sentiment Classification &  NLI \\ \Xhline{2\arrayrulewidth}
                                              
    \multirow{4}{*}{16}   & \method$_{\mbox{10}}$  & 37.62 \tiny{$\pm$ 7.37} &  58.10 \tiny{$\pm$ 5.40} & 78.53 \tiny{$\pm$ 1.55}   \\
                          & \method$_{\mbox{5}}$ &  62.49 \tiny{$\pm$ 4.23} &  71.50 \tiny{$\pm$ 5.93} &  73.27 \tiny{$\pm$ 2.63} \\
                          & \method &  69.00 \tiny{$\pm$ 4.76}  &  76.65 \tiny{$\pm$ 2.47} &  76.10 \tiny{$\pm$ 2.21} \\
                          & \method-ZERO & 44.79 \tiny{$\pm$ 9.34}    & 74.45 \tiny{$\pm$ 3.34}  & 74.36 \tiny{$\pm$ 6.67}  \\
\Xhline{2\arrayrulewidth}
\end{tabular}
\caption{Ablations: $\mbox{\method}_{\nu}$ does not adapt layers $0-\nu$ (inclusive) in the inner loop (and fine-tuning), while \method adapts all parameters. Note that the outer loop still optimizes all parameters. For new tasks (like entity typing) adapting all parameters is better while for tasks seen at training time (like NLI) adapting fewer parameters is better.
\method-ZERO is a model trained without the softmax-generator and a zero initialized softmax classifier, which shows the importance of softmax generator in \method.}
\label{tab:efficient}
\end{table*}

For ablations we use the dev-set of 3 tasks: CoNLL-2003 entity typing,  Amazon reviews DVD domain sentiment classification and SciTail NLI. 

\noindent\textbf{Importance of softmax parameters}:
Since the softmax generation is an important component of \method, we study how it affects performance. We remove the softmax generator and instead add a softmax weight and bias with zero initialization for each task. The model is trained in a similar way as \method. This method, termed \method-ZERO, is a naive application of MAML to this problem. Table \ref{tab:efficient} shows that this performs worse on new tasks, highlighting the importance of softmax generator.

\noindent\textbf{Parameter efficiency}: We consider three variants of \method with parameter efficient training discussed in Sec~\ref{sec:adapt}.
Denote $\mbox{\method}_\nu$ as the model which does not adapt layers 0 to $\nu$ (including word embeddings) in the inner loop of meta-training.
Note that even for $\nu \ne 0$, the parameters are still optimized in the outer loop.
Table \ref{tab:efficient} shows the results.
Interestingly, for all tasks (except NLI) we find that adapting all parameters is better. 
This is potentially because the per-layer learning rate in \method also adjust the adaptation rates for each layer.
On SciTail (NLI) we observe the opposite behaviour, suggesting that adapting fewer parameters is better for small $k$, potentially because training consisted of multiple NLI datasets.

\begin{figure*}[t!]
  \centering
  \includegraphics[width=\linewidth]{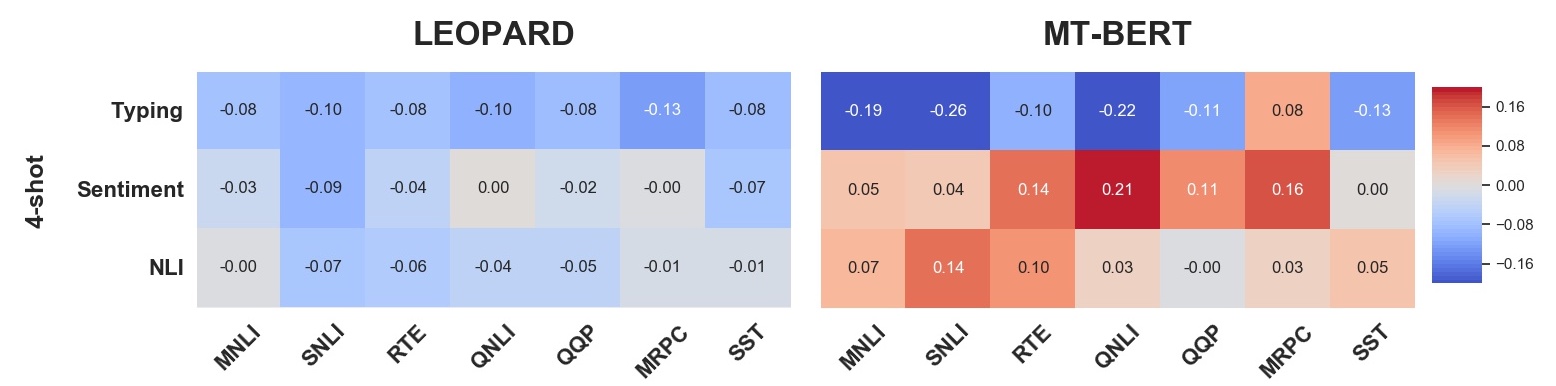}
  \caption{Analyzing target task performance as a function of training tasks (best viewed in color). 
  Each column represents one held-out training task (name on $x$-axis) and each row corresponds to one target task (name on $y$-axis). 
  Each cell is the relative change in performance on the target task when the corresponding training task is held-out, compared to training on all the train tasks.
  Dark blue indicates large drop, dark red indicates large increase and grey indicates close to no change in performance.
  In general, \method's performance is more consistent compared to MT-BERT indicating that meta-training learns more generalized initial parameters compared to multi-task training.}
  \label{fig:heldout}
\end{figure*}

\noindent\textbf{Importance of training tasks}:
We study how target-task performance of MT-BERT and \method is dependent on tasks used for training. 
For this experiment, we held out each training task one by one and trained both models.
The trained models are then evaluated for their performance on the target tasks (using the development set), following the same protocol as before.
Fig.~\ref{fig:heldout} shows a visualization of the relative change in performance when each training task is held out. 
We see that \method's performance is more consistent with respect to variation in training tasks, owing
to the meta-training procedure that finds an initial point that performs equally well across tasks.
Removing a task often leads to decrease in performance for \method as it decreases the number of meta-training tasks and leads to over-fitting to the training task-distribution. 
In contrast, MT-BERT's performance on target tasks varies greatly depending on the held-in training tasks. 

\section{Related Work}
\label{sec:related}
Meta-Learning approaches can be broadly classified as: optimization-based \cite{Finn:2017:MMF:3305381.3305498,al2017continuous,nichol2018reptile,rusu2018metalearning}, 
model-based 
\cite{santoro2016meta,ravi2017optimization,munkhdalai2017meta},
and metric-learning based
\cite{vinyals2016matching,snell2017prototypical,sung2018learning}.
We refer to \newcite{finn2018learning} for an exhaustive review.
Recently, it has been shown that learning task-dependent model parameters improves few-shot learning \cite{rusu2018metalearning,zintgraf2018cavia}. 
While existing methods train and evaluate on simulated datasets with limited diversity, there is recent interest for more realistic meta-learning applications \cite{triantafillou2019metadataset} and our work significantly advances this by training and evaluating on diverse and real NLP tasks.

Meta-learning applications in NLP have yielded improvements on specific tasks.
\newcite{gu2018meta} used MAML to simulate low resource machine translation,
\newcite{chen2018meta} learn HyperLSTM \cite{ha2016hypernetworks} model in a multi-task setting across various sentiment classification domains,
and other recent approaches \cite{guo2018multi,yu2018diverse,han2018fewrel,obamuyide2019model,geng2019induction,mi2019meta,DBLP:conf/iclr/BaoWCB20} meta-train for a \textit{specific} classification task, such as relation classification,
and do not generalize beyond the training task.
\newcite{dou2019investigating} train on a subset of GLUE tasks to generalize to other GLUE tasks and their approach does not consider unseen tasks.
Transfer learning is a closely related research area. 
Self-supervised pre-training has been shown to learn general-purpose model parameters that improve downstream performance with fine-tuning 
\cite{peters2018deep,howard2018universal,devlin2018bert,radford2019language,yang2019xlnet,raffel2019exploring}. 
Fine-tuning, however, typically requires large training data \cite{yogatama2018learning}. 
Multi-task learning with BERT has been shown to improve performance for many related tasks \cite{phang2018sentence,wang2018can,liu2019multi}.
We refer the reader to \newcite{ruder2019neural} for a more thorough discussion of transfer learning and multi-task learning.

\section{Conclusions}
Learning general linguistic intelligence has been a long-term goal of NLP.
While humans, with all their prior knowledge, can quickly learn to solve new tasks with very few examples, machine-learned models still struggle to demonstrate such intelligence.
To this end, we proposed \method, a meta-learning approach, and found that it learns more general-purpose parameters that better prime the model to solve completely new tasks with few examples.
While we see improvements using meta-learning, performance with few examples still lags behind human-level performance.
We consider bridging this gap as a lucrative goal to demonstrate general linguistic intelligence, and meta-learning as a strong contender to achieve this goal.

\section{Acknowledgements}
We will like to thank Kalpesh Krishna, Tu Vu and Tsendsuren Munkhdalai for feedback on earlier drafts of this manuscript.
This work was supported in part by the Chan Zuckerberg Initiative, and in part by the National Science Foundation under Grant No. IIS-1514053 and IIS-1763618. Any opinions, findings and conclusions or recommendations expressed in this material are those of the authors and do not necessarily reflect those of the sponsor.

\bibliographystyle{coling}
\bibliography{references}

\begin{thebibliography}{}

\bibitem[\protect\citename{Al-Shedivat \bgroup et al.\egroup
  }2018]{al2017continuous}
Maruan Al-Shedivat, Trapit Bansal, Yuri Burda, Ilya Sutskever, Igor Mordatch,
  and Pieter Abbeel.
\newblock 2018.
\newblock Continuous adaptation via meta-learning in nonstationary and
  competitive environments.
\newblock In {\em Proceedings of the International Conference on Learning
  Representations}.

\bibitem[\protect\citename{Antoniou \bgroup et al.\egroup
  }2018]{antoniou2018train}
Antreas Antoniou, Harrison Edwards, and Amos Storkey.
\newblock 2018.
\newblock How to train your maml.
\newblock {\em arXiv preprint arXiv:1810.09502}.

\bibitem[\protect\citename{Ba \bgroup et al.\egroup }2016]{ba2016layer}
Jimmy~Lei Ba, Jamie~Ryan Kiros, and Geoffrey~E Hinton.
\newblock 2016.
\newblock Layer normalization.
\newblock {\em arXiv preprint arXiv:1607.06450}.

\bibitem[\protect\citename{Bao \bgroup et al.\egroup
  }2020]{DBLP:conf/iclr/BaoWCB20}
Yujia Bao, Menghua Wu, Shiyu Chang, and Regina Barzilay.
\newblock 2020.
\newblock Few-shot text classification with distributional signatures.
\newblock In {\em 8th International Conference on Learning Representations,
  {ICLR} 2020, Addis Ababa, Ethiopia, April 26-30, 2020}. OpenReview.net.

\bibitem[\protect\citename{Bengio \bgroup et al.\egroup
  }1992]{bengio1992optimization}
Samy Bengio, Yoshua Bengio, Jocelyn Cloutier, and Jan Gecsei.
\newblock 1992.
\newblock On the optimization of a synaptic learning rule.
\newblock In {\em Preprints Conf. Optimality in Artificial and Biological
  Neural Networks}, pages 6--8. Univ. of Texas.

\bibitem[\protect\citename{Blitzer \bgroup et al.\egroup
  }2007]{blitzer2007biographies}
John Blitzer, Mark Dredze, and Fernando Pereira.
\newblock 2007.
\newblock Biographies, bollywood, boom-boxes and blenders: Domain adaptation
  for sentiment classification.
\newblock In {\em Proceedings of the 45th annual meeting of the association of
  computational linguistics}, pages 440--447.

\bibitem[\protect\citename{Bowman \bgroup et al.\egroup }2015]{bowman2015snli}
Samuel~R Bowman, Gabor Angeli, Christopher Potts, and Christopher~D Manning.
\newblock 2015.
\newblock A large annotated corpus for learning natural language inference.
\newblock In {\em Proceedings of the 2015 Conference on Empirical Methods in
  Natural Language Processing}, pages 632--642.

\bibitem[\protect\citename{Caruana}1997]{caruana1997multitask}
Rich Caruana.
\newblock 1997.
\newblock Multitask learning.
\newblock {\em Machine learning}, 28(1):41--75.

\bibitem[\protect\citename{Chen \bgroup et al.\egroup }2018]{chen2018meta}
Junkun Chen, Xipeng Qiu, Pengfei Liu, and Xuanjing Huang.
\newblock 2018.
\newblock Meta multi-task learning for sequence modeling.
\newblock In {\em Thirty-Second AAAI Conference on Artificial Intelligence}.

\bibitem[\protect\citename{Dagan \bgroup et al.\egroup }2005]{dagan2005pascal}
Ido Dagan, Oren Glickman, and Bernardo Magnini.
\newblock 2005.
\newblock The pascal recognising textual entailment challenge.
\newblock In {\em Machine Learning Challenges Workshop}, pages 177--190.
  Springer.

\bibitem[\protect\citename{Devlin \bgroup et al.\egroup }2018]{devlin2018bert}
Jacob Devlin, Ming-Wei Chang, Kenton Lee, and Kristina Toutanova.
\newblock 2018.
\newblock Bert: Pre-training of deep bidirectional transformers for language
  understanding.
\newblock {\em arXiv preprint arXiv:1810.04805}.

\bibitem[\protect\citename{Dolan and Brockett}2005]{dolan2005automatically}
William~B Dolan and Chris Brockett.
\newblock 2005.
\newblock Automatically constructing a corpus of sentential paraphrases.
\newblock In {\em Proceedings of the Third International Workshop on
  Paraphrasing (IWP2005)}.

\bibitem[\protect\citename{Dou \bgroup et al.\egroup
  }2019]{dou2019investigating}
Zi-Yi Dou, Keyi Yu, and Antonios Anastasopoulos.
\newblock 2019.
\newblock Investigating meta-learning algorithms for low-resource natural
  language understanding tasks.
\newblock In {\em Proceedings of the 2019 Conference on Empirical Methods in
  Natural Language Processing and the 9th International Joint Conference on
  Natural Language Processing (EMNLP-IJCNLP)}, pages 1192--1197.

\bibitem[\protect\citename{Finn \bgroup et al.\egroup
  }2017]{Finn:2017:MMF:3305381.3305498}
Chelsea Finn, Pieter Abbeel, and Sergey Levine.
\newblock 2017.
\newblock Model-agnostic meta-learning for fast adaptation of deep networks.
\newblock In {\em Proceedings of the 34th International Conference on Machine
  Learning - Volume 70}, pages 1126--1135.

\bibitem[\protect\citename{Finn}2018]{finn2018learning}
Chelsea Finn.
\newblock 2018.
\newblock {\em Learning to Learn with Gradients}.
\newblock {Ph.D.} thesis, UC Berkeley.

\bibitem[\protect\citename{Geng \bgroup et al.\egroup }2019]{geng2019induction}
Ruiying Geng, Binhua Li, Yongbin Li, Xiaodan Zhu, Ping Jian, and Jian Sun.
\newblock 2019.
\newblock Induction networks for few-shot text classification.

\bibitem[\protect\citename{Giampiccolo \bgroup et al.\egroup
  }2007]{giampiccolo2007third}
Danilo Giampiccolo, Bernardo Magnini, Ido Dagan, and Bill Dolan.
\newblock 2007.
\newblock The third pascal recognizing textual entailment challenge.
\newblock In {\em Proceedings of the ACL-PASCAL workshop on textual entailment
  and paraphrasing}, pages 1--9.

\bibitem[\protect\citename{Giampiccolo \bgroup et al.\egroup
  }2008]{giampiccolo2008fourth}
Danilo Giampiccolo, Hoa~Trang Dang, Bernardo Magnini, Ido Dagan, Elena Cabrio,
  and Bill Dolan.
\newblock 2008.
\newblock The fourth pascal recognizing textual entailment challenge.
\newblock In {\em TAC}. Citeseer.

\bibitem[\protect\citename{Gu \bgroup et al.\egroup }2018]{gu2018meta}
Jiatao Gu, Yong Wang, Yun Chen, Kyunghyun Cho, and Victor~OK Li.
\newblock 2018.
\newblock Meta-learning for low-resource neural machine translation.
\newblock {\em arXiv preprint arXiv:1808.08437}.

\bibitem[\protect\citename{Guo \bgroup et al.\egroup }2018]{guo2018multi}
Jiang Guo, Darsh Shah, and Regina Barzilay.
\newblock 2018.
\newblock Multi-source domain adaptation with mixture of experts.
\newblock In {\em Proceedings of the 2018 Conference on Empirical Methods in
  Natural Language Processing}, pages 4694--4703.

\bibitem[\protect\citename{Ha \bgroup et al.\egroup }2016]{ha2016hypernetworks}
David Ha, Andrew Dai, and Quoc~V Le.
\newblock 2016.
\newblock Hypernetworks.
\newblock {\em arXiv preprint arXiv:1609.09106}.

\bibitem[\protect\citename{Haim \bgroup et al.\egroup }2006]{haim2006second}
R~Bar Haim, Ido Dagan, Bill Dolan, Lisa Ferro, Danilo Giampiccolo, Bernardo
  Magnini, and Idan Szpektor.
\newblock 2006.
\newblock The second pascal recognising textual entailment challenge.
\newblock In {\em Proceedings of the Second PASCAL Challenges Workshop on
  Recognising Textual Entailment}.

\bibitem[\protect\citename{Han \bgroup et al.\egroup }2018]{han2018fewrel}
Xu~Han, Hao Zhu, Pengfei Yu, Ziyun Wang, Yuan Yao, Zhiyuan Liu, and Maosong
  Sun.
\newblock 2018.
\newblock Fewrel: A large-scale supervised few-shot relation classification
  dataset with state-of-the-art evaluation.
\newblock In {\em Proceedings of the 2018 Conference on Empirical Methods in
  Natural Language Processing}, pages 4803--4809.

\bibitem[\protect\citename{Houlsby \bgroup et al.\egroup
  }2019]{houlsby2019parameter}
Neil Houlsby, Andrei Giurgiu, Stanislaw Jastrzebski, Bruna Morrone, Quentin
  De~Laroussilhe, Andrea Gesmundo, Mona Attariyan, and Sylvain Gelly.
\newblock 2019.
\newblock Parameter-efficient transfer learning for nlp.
\newblock In {\em International Conference on Machine Learning}.

\bibitem[\protect\citename{Howard and Ruder}2018]{howard2018universal}
Jeremy Howard and Sebastian Ruder.
\newblock 2018.
\newblock Universal language model fine-tuning for text classification.
\newblock In {\em Proceedings of the 56th Annual Meeting of the Association for
  Computational Linguistics (Volume 1: Long Papers)}, pages 328--339.

\bibitem[\protect\citename{Kann \bgroup et al.\egroup }2019]{kann2019towards}
Katharina Kann, Kyunghyun Cho, and Samuel~R Bowman.
\newblock 2019.
\newblock Towards realistic practices in low-resource natural language
  processing: The development set.
\newblock {\em arXiv preprint arXiv:1909.01522}.

\bibitem[\protect\citename{Khot \bgroup et al.\egroup }2018]{khot2018scitail}
Tushar Khot, Ashish Sabharwal, and Peter Clark.
\newblock 2018.
\newblock Scitail: A textual entailment dataset from science question
  answering.
\newblock In {\em Thirty-Second AAAI Conference on Artificial Intelligence}.

\bibitem[\protect\citename{Li \bgroup et al.\egroup }2017]{li2017meta}
Zhenguo Li, Fengwei Zhou, Fei Chen, and Hang Li.
\newblock 2017.
\newblock Meta-sgd: Learning to learn quickly for few-shot learning.
\newblock {\em arXiv preprint arXiv:1707.09835}.

\bibitem[\protect\citename{Liu \bgroup et al.\egroup }2013]{liu2013asgard}
Jingjing Liu, Panupong Pasupat, Scott Cyphers, and Jim Glass.
\newblock 2013.
\newblock Asgard: A portable architecture for multilingual dialogue systems.
\newblock In {\em 2013 IEEE International Conference on Acoustics, Speech and
  Signal Processing}, pages 8386--8390. IEEE.

\bibitem[\protect\citename{Liu \bgroup et al.\egroup }2019]{liu2019multi}
Xiaodong Liu, Pengcheng He, Weizhu Chen, and Jianfeng Gao.
\newblock 2019.
\newblock Multi-task deep neural networks for natural language understanding.
\newblock {\em arXiv preprint arXiv:1901.11504}.

\bibitem[\protect\citename{Mi \bgroup et al.\egroup }2019]{mi2019meta}
Fei Mi, Minlie Huang, Jiyong Zhang, and Boi Faltings.
\newblock 2019.
\newblock Meta-learning for low-resource natural language generation in
  task-oriented dialogue systems.
\newblock {\em arXiv preprint arXiv:1905.05644}.

\bibitem[\protect\citename{Munkhdalai and Yu}2017]{munkhdalai2017meta}
Tsendsuren Munkhdalai and Hong Yu.
\newblock 2017.
\newblock Meta networks.
\newblock In {\em Proceedings of the 34th International Conference on Machine
  Learning-Volume 70}, pages 2554--2563. JMLR. org.

\bibitem[\protect\citename{Nichol and Schulman}2018]{nichol2018reptile}
Alex Nichol and John Schulman.
\newblock 2018.
\newblock Reptile: a scalable metalearning algorithm.
\newblock {\em arXiv preprint arXiv:1803.02999}, 2.

\bibitem[\protect\citename{Obamuyide and Vlachos}2019]{obamuyide2019model}
Abiola Obamuyide and Andreas Vlachos.
\newblock 2019.
\newblock Model-agnostic meta-learning for relation classification with limited
  supervision.
\newblock In {\em Proceedings of the 57th Annual Meeting of the Association for
  Computational Linguistics}, pages 5873--5879.

\bibitem[\protect\citename{Peters \bgroup et al.\egroup }2018]{peters2018deep}
Matthew~E Peters, Mark Neumann, Mohit Iyyer, Matt Gardner, Christopher Clark,
  Kenton Lee, and Luke Zettlemoyer.
\newblock 2018.
\newblock Deep contextualized word representations.
\newblock In {\em Proceedings of NAACL-HLT}, pages 2227--2237.

\bibitem[\protect\citename{Phang \bgroup et al.\egroup
  }2018]{phang2018sentence}
Jason Phang, Thibault F{\'e}vry, and Samuel~R Bowman.
\newblock 2018.
\newblock Sentence encoders on stilts: Supplementary training on intermediate
  labeled-data tasks.
\newblock {\em arXiv preprint arXiv:1811.01088}.

\bibitem[\protect\citename{Radford \bgroup et al.\egroup
  }2019]{radford2019language}
Alec Radford, Jeffrey Wu, Rewon Child, David Luan, Dario Amodei, and Ilya
  Sutskever.
\newblock 2019.
\newblock Language models are unsupervised multitask learners.
\newblock {\em OpenAI Blog}, 1(8).

\bibitem[\protect\citename{Raffel \bgroup et al.\egroup
  }2019]{raffel2019exploring}
Colin Raffel, Noam Shazeer, Adam Roberts, Katherine Lee, Sharan Narang, Michael
  Matena, Yanqi Zhou, Wei Li, and Peter~J Liu.
\newblock 2019.
\newblock Exploring the limits of transfer learning with a unified text-to-text
  transformer.
\newblock {\em arXiv preprint arXiv:1910.10683}.

\bibitem[\protect\citename{Rajpurkar \bgroup et al.\egroup
  }2016]{rajpurkar2016squad}
Pranav Rajpurkar, Jian Zhang, Konstantin Lopyrev, and Percy Liang.
\newblock 2016.
\newblock Squad: 100,000+ questions for machine comprehension of text.
\newblock {\em arXiv preprint arXiv:1606.05250}.

\bibitem[\protect\citename{Ravi and Larochelle}2017]{ravi2017optimization}
Sachin Ravi and Hugo Larochelle.
\newblock 2017.
\newblock Optimization as a model for few-shot learning.
\newblock In {\em Proceedings of the International Conference on Learning
  Representations}.

\bibitem[\protect\citename{Ruder}2019]{ruder2019neural}
Sebastian Ruder.
\newblock 2019.
\newblock {\em Neural Transfer Learning for Natural Language Processing}.
\newblock {Ph.D.} thesis, NATIONAL UNIVERSITY OF IRELAND, GALWAY.

\bibitem[\protect\citename{Rusu \bgroup et al.\egroup
  }2019]{rusu2018metalearning}
Andrei~A. Rusu, Dushyant Rao, Jakub Sygnowski, Oriol Vinyals, Razvan Pascanu,
  Simon Osindero, and Raia Hadsell.
\newblock 2019.
\newblock Meta-learning with latent embedding optimization.
\newblock In {\em International Conference on Learning Representations}.

\bibitem[\protect\citename{Sang and De~Meulder}2003]{sang2003conll}
Erik F Tjong~Kim Sang and Fien De~Meulder.
\newblock 2003.
\newblock Introduction to the conll-2003 shared task: Language-independent
  named entity recognition.
\newblock In {\em Proceedings of the Seventh Conference on Natural Language
  Learning at HLT-NAACL 2003}, pages 142--147.

\bibitem[\protect\citename{Santoro \bgroup et al.\egroup
  }2016]{santoro2016meta}
Adam Santoro, Sergey Bartunov, Matthew Botvinick, Daan Wierstra, and Timothy
  Lillicrap.
\newblock 2016.
\newblock Meta-learning with memory-augmented neural networks.
\newblock In {\em International conference on machine learning}, pages
  1842--1850.

\bibitem[\protect\citename{Schmidhuber}1987]{schmidhuber1987evolutionary}
J{\"u}rgen Schmidhuber.
\newblock 1987.
\newblock {\em Evolutionary principles in self-referential learning, or on
  learning how to learn: the meta-meta-... hook}.
\newblock {Ph.D.} thesis, Technische Universit{\"a}t M{\"u}nchen.

\bibitem[\protect\citename{Snell \bgroup et al.\egroup
  }2017]{snell2017prototypical}
Jake Snell, Kevin Swersky, and Richard Zemel.
\newblock 2017.
\newblock Prototypical networks for few-shot learning.
\newblock In {\em Advances in Neural Information Processing Systems}, pages
  4077--4087.

\bibitem[\protect\citename{Socher \bgroup et al.\egroup
  }2013]{socher2013recursive}
Richard Socher, Alex Perelygin, Jean Wu, Jason Chuang, Christopher~D Manning,
  Andrew Ng, and Christopher Potts.
\newblock 2013.
\newblock Recursive deep models for semantic compositionality over a sentiment
  treebank.
\newblock In {\em Proceedings of the 2013 conference on empirical methods in
  natural language processing}, pages 1631--1642.

\bibitem[\protect\citename{Sung \bgroup et al.\egroup }2018]{sung2018learning}
Flood Sung, Yongxin Yang, Li~Zhang, Tao Xiang, Philip~HS Torr, and Timothy~M
  Hospedales.
\newblock 2018.
\newblock Learning to compare: Relation network for few-shot learning.
\newblock In {\em Proceedings of the IEEE Conference on Computer Vision and
  Pattern Recognition}, pages 1199--1208.

\bibitem[\protect\citename{Thrun and Pratt}2012]{thrun2012learning}
Sebastian Thrun and Lorien Pratt.
\newblock 2012.
\newblock {\em Learning to learn}.
\newblock Springer Science \& Business Media.

\bibitem[\protect\citename{Triantafillou \bgroup et al.\egroup
  }2019]{triantafillou2019metadataset}
Eleni Triantafillou, Tyler Zhu, Vincent Dumoulin, Pascal Lamblin, Utku Evci,
  Kelvin Xu, Ross Goroshin, Carles Gelada, Kevin Swersky, Pierre-Antoine
  Manzagol, and Hugo Larochelle.
\newblock 2019.
\newblock Meta-dataset: A dataset of datasets for learning to learn from few
  examples.

\bibitem[\protect\citename{Vaswani \bgroup et al.\egroup
  }2017]{vaswani2017attention}
Ashish Vaswani, Noam Shazeer, Niki Parmar, Jakob Uszkoreit, Llion Jones,
  Aidan~N Gomez, {\L}ukasz Kaiser, and Illia Polosukhin.
\newblock 2017.
\newblock Attention is all you need.
\newblock In {\em Advances in neural information processing systems}, pages
  5998--6008.

\bibitem[\protect\citename{Vinyals \bgroup et al.\egroup
  }2016]{vinyals2016matching}
Oriol Vinyals, Charles Blundell, Timothy Lillicrap, Daan Wierstra, et~al.
\newblock 2016.
\newblock Matching networks for one shot learning.
\newblock In {\em Advances in neural information processing systems}, pages
  3630--3638.

\bibitem[\protect\citename{Wang \bgroup et al.\egroup }2018a]{wang2018can}
Alex Wang, Jan Hula, Patrick Xia, Raghavendra Pappagari, R~Thomas McCoy, Roma
  Patel, Najoung Kim, Ian Tenney, Yinghui Huang, Katherin Yu, et~al.
\newblock 2018a.
\newblock Can you tell me how to get past sesame street? sentence-level
  pretraining beyond language modeling.
\newblock {\em arXiv preprint arXiv:1812.10860}.

\bibitem[\protect\citename{Wang \bgroup et al.\egroup }2018b]{wang2018glue}
Alex Wang, Amanpreet Singh, Julian Michael, Felix Hill, Omer Levy, and Samuel~R
  Bowman.
\newblock 2018b.
\newblock Glue: A multi-task benchmark and analysis platform for natural
  language understanding.
\newblock {\em arXiv preprint arXiv:1804.07461}.

\bibitem[\protect\citename{Warstadt \bgroup et al.\egroup
  }2019]{warstadt2019neural}
Alex Warstadt, Amanpreet Singh, and Samuel~R Bowman.
\newblock 2019.
\newblock Neural network acceptability judgments.
\newblock {\em Transactions of the Association for Computational Linguistics},
  7:625--641.

\bibitem[\protect\citename{Williams \bgroup et al.\egroup
  }2017]{williams2017broad}
Adina Williams, Nikita Nangia, and Samuel~R Bowman.
\newblock 2017.
\newblock A broad-coverage challenge corpus for sentence understanding through
  inference.
\newblock {\em arXiv preprint arXiv:1704.05426}.

\bibitem[\protect\citename{Yang \bgroup et al.\egroup }2019]{yang2019xlnet}
Zhilin Yang, Zihang Dai, Yiming Yang, Jaime Carbonell, Ruslan Salakhutdinov,
  and Quoc~V Le.
\newblock 2019.
\newblock Xlnet: Generalized autoregressive pretraining for language
  understanding.
\newblock {\em arXiv preprint arXiv:1906.08237}.

\bibitem[\protect\citename{Yogatama \bgroup et al.\egroup
  }2019]{yogatama2018learning}
Dani Yogatama, Cyprien de~Masson~d'Autume, Jerome Connor, Tom{\'{a}}s
  Kocisk{\'{y}}, Mike Chrzanowski, Lingpeng Kong, Angeliki Lazaridou, Wang
  Ling, Lei Yu, Chris Dyer, and Phil Blunsom.
\newblock 2019.
\newblock Learning and evaluating general linguistic intelligence.
\newblock {\em CoRR}, abs/1901.11373.

\bibitem[\protect\citename{Yu \bgroup et al.\egroup }2018]{yu2018diverse}
Mo~Yu, Xiaoxiao Guo, Jinfeng Yi, Shiyu Chang, Saloni Potdar, Yu~Cheng, Gerald
  Tesauro, Haoyu Wang, and Bowen Zhou.
\newblock 2018.
\newblock Diverse few-shot text classification with multiple metrics.
\newblock {\em arXiv preprint arXiv:1805.07513}.

\bibitem[\protect\citename{Zintgraf \bgroup et al.\egroup
  }2019]{zintgraf2018cavia}
Luisa~M Zintgraf, Kyriacos Shiarlis, Vitaly Kurin, Katja Hofmann, and Shimon
  Whiteson.
\newblock 2019.
\newblock Cavia: Fast context adaptation via meta-learning.
\newblock In {\em International Conference on Machine Learning}.

\end{thebibliography}

\appendix
\section{Appendix}
\begin{appendix}


\section{Datasets}\label{sec:moredata}
\noindent\textbf{Data Augmentation}:
Meta-learning benefits from training across many tasks. 
We thus create multiple versions of tasks with more than 2 classes by considering classifying between every pair of labels as a task. 
Existing methods \cite{vinyals2016matching,snell2017prototypical,Finn:2017:MMF:3305381.3305498} treat each random sample of labels from a pool of labels (for example in image classification) as a task. 
In order to create more diversity during training, we also create multiple versions of each dataset that has more than 2 classes, by considering classifying between every possible pair of labels as a training task. 
This increases the number of tasks and allows for more per-label examples in a batch during training. 
In addition, since one of the goals is to learn to classify phrases in a sentence, we modify the sentiment classification task (SST-2) in GLUE, which contains annotations of sentiment for phrases, by providing a sentence in which the phrase occurs as part of the input. That is, the input is the sentence followed by a separator token \cite{devlin2018bert} followed by the phrase to classify. 
An example of the input to all the models for the entity typing tasks can be found in Table \ref{tab:typing_ex}
\begin{table*}[htb]
\centering \fontsize{8.8}{10.5}\selectfont \setlength{\tabcolsep}{0.5em}
\begin{tabular}{cc}
Input & Label \\ \hline
are there any [authentic mexican]$^1$ restaurants in [the area]$^2$ & $^1$Cuisine, $^2$Location\\ \hline
are there any authentic mexican restaurants in the area [SEP] authentic mexican & Cuisine\\
are there any authentic mexican restaurants in the area [SEP] the area & Location \\
\end{tabular}
\caption{An example of an input from the MIT restaurants dataset. The first line is the actual example with two mentions. The next two lines are the input to the models -- one for each mention.}
\label{tab:typing_ex}
\end{table*}

We use the standard train, dev data split for GLUE and SNLI \cite{wang2018glue,bowman2015snli}. 
For our ablation studies, on our target task we take 20\% of the training data as validation for early stopping and sample from the remaining 80\% to create the few-shot data for fine-tuning. 
For training MT-BERT we use dev data of the training task as the validation set. For meta-learning methods, prototypical network and \method, we use additional validation datasets as is typical in meta learning \cite{Finn:2017:MMF:3305381.3305498,snell2017prototypical}. We use unlabelled Amazon review data from apparel, health, software, toys, video as categorization tasks and labelled data from music, toys, video as sentiment classification task. 

Details of the datasets are present in Table \ref{tab:dataset}.
\begin{table*}[ht]
\centering
\resizebox{\linewidth}{!}{%
\begin{tabular}{ccccccc}
\Xhline{2\arrayrulewidth}
Dataset & Labels & Training Size & Validation Size &Testing Size  & Source\\ \hline

ARSC Domains & 2 & 800 & 200 & 1000 & \cite{blitzer2007biographies} \\ \hline
CoLA & 2 & 8551 & 1042 & --- & \cite{warstadt2019neural} \\ \hline
MRPC & 2 & 3669 & 409 & --- & \cite{dolan2005automatically} \\ \hline
QNLI & 2 & 104744 & 5464 & ---  & \cite{rajpurkar2016squad,wang2018glue} \\ \hline
QQP & 2 & 363847 & 40431 & ---  & \cite{wang2018glue} \\ \hline
RTE & 2 & 2491 & 278 & ---  & \cite{dagan2005pascal,haim2006second,giampiccolo2007third,giampiccolo2008fourth} \\ \hline
SNLI & 3 & 549368 & 9843 & ---  & \cite{bowman2015snli} \\ \hline
SST-2 & 2 & 67350 & 873 & ---  & \cite{socher2013recursive} \\ \hline
MNLI \tiny{(m/mm)} & 3 & 392703 & 19649 & ---  & \cite{williams2017broad} \\ \hline
Scitail & 2 & 23,596 & 1,304 & 2,126  & \cite{khot2018scitail} \\ \hline
Airline & 3 & 7320 & --- & 7320  & \url{https://www.figure-eight.com/data-for-everyone/} \\ \hline
Disaster & 2 & 4887 & --- & 4887  &  \url{https://www.figure-eight.com/data-for-everyone/} \\ \hline
Political Bias & 2 & 2500 & --- & 2500  & \url{https://www.figure-eight.com/data-for-everyone/} \\ \hline
Political Audience & 2 & 2500 & --- & 2500  & \url{https://www.figure-eight.com/data-for-everyone/}\\ \hline
Political Message & 9 & 2500 & --- & 2500  & \url{https://www.figure-eight.com/data-for-everyone/} \\ \hline
Emotion & 13 & 20000 & --- & 20000  & \url{https://www.figure-eight.com/data-for-everyone/} \\ \hline
CoNLL & 4 & 23499 & 5942 & 5648  & \cite{sang2003conll} \\ \hline
MIT-Restaurant & 8 & 12474 & --- & 2591  & \cite{liu2013asgard} \url{https://groups.csail.mit.edu/sls/downloads/restaurant/} \\ \hline
\end{tabular}
}
\caption{Dataset statistics for all the datasets used in our analysis. "-" represent data that is either not available or not used in this study. We have balanced severely unbalanced datasets(Political Bias and Audience) as our training data is balanced. To create training data for few shot experiments we sample 10 datasets for each k-shot. *Sec \ref{sec:moredata} for more details}
\label{tab:dataset}
\end{table*}

\subsection{Test Datasets}
The tasks and datasets we used for evaluating performance on few-shot learning are as follows:
\begin{enumerate}
    \item Entity Typing: We use the following datasets for entity typing: CoNLL-2003 \cite{sang2003conll} and MIT-Restaurant \cite{liu2013asgard}. Note that we consider each mention as a separate labelled example. CoNLL dataset consists of text from news articles while MIT dataset contains text from restaurant queries.
    \item Sentiment Classification: We use the sentiment annotated data from Amazon Reviews dataset \cite{blitzer2007biographies} which contains user reviews and the binary sentiment for various domains of products. We use the Books, DVD, Electronics, and Kitchen \& Housewares domains, which are commonly used domains in the literature \cite{yu2018diverse}. 
    \item Rating Classification: We use the ratings from the Amazon Reviews dataset \cite{blitzer2007biographies} which is not annotated with overall sentiment, and consider classifying into 3 classes: rating $\le 2$, rating $=4$ and rating $=5$.
    \item Text Classification: We use multiple text classification datasets from crowdflower\footnote{https://www.figure-eight.com/data-for-everyone/}. These involve classifying sentiments of tweets towards an airline, classifying whether a tweet refers to a disaster event, classifying emotional content of text, classifying the audience/bias/message of social media messages from politicians. These tasks are quite different from the training tasks both in terms of the labels as well as the input domain. 
    \item NLI: We use the SciTail dataset \cite{khot2018scitail}, which is a dataset for entailment created from science questions.
\end{enumerate}

\section{Additional Results}
\label{sec:more_results}

\begin{table*}[htb!]
\centering \fontsize{8.8}{10.5}\selectfont \setlength{\tabcolsep}{0.5em}
\begin{tabular}{cccccccc}
\Xhline{2\arrayrulewidth}
\multicolumn{8}{c}{\textbf{Amazon Review Sentiment Classification}}    \\ \Xhline{2\arrayrulewidth}
\multicolumn{2}{l}{} & $\mbox{BERT}_{\mbox{base}}$ & $\mbox{MT-BERT}_{\mbox{softmax}}$ & MT-BERT & $\mbox{MT-BERT}_{\mbox{resue}}$ & Proto-BERT & \method \\[3pt] \Xhline{2\arrayrulewidth}
\multirow{4}{*}{Books} & 4 &54.81 \tiny{$\pm$ 3.75} & 68.69 \tiny{$\pm$ 5.21} & 64.93 \tiny{$\pm$ 8.65} & 74.79 \tiny{$\pm$ 6.91} & 73.15 \tiny{$\pm$ 5.85} & 82.54 \tiny{$\pm$ 1.33} \\
 & 8 &53.54 \tiny{$\pm$ 5.17} & 74.86 \tiny{$\pm$ 2.17} & 67.38 \tiny{$\pm$ 9.78} & 78.21 \tiny{$\pm$ 3.49} & 75.46 \tiny{$\pm$ 6.87} & 83.03 \tiny{$\pm$ 1.28} \\
 & 16 &65.56 \tiny{$\pm$ 4.12} & 74.88 \tiny{$\pm$ 4.34} & 69.65 \tiny{$\pm$ 8.94} & 78.87 \tiny{$\pm$ 3.32} & 77.26 \tiny{$\pm$ 3.27} & 83.33 \tiny{$\pm$ 0.79} \\
                    \midrule
\multirow{4}{*}{DVD} & 4 &54.98 \tiny{$\pm$ 3.96} & 63.68 \tiny{$\pm$ 5.03} & 66.36 \tiny{$\pm$ 7.46} & 71.74 \tiny{$\pm$ 8.54} & 74.38 \tiny{$\pm$ 2.44} & 80.32 \tiny{$\pm$ 1.02} \\
 & 8 &55.63 \tiny{$\pm$ 4.34} & 67.54 \tiny{$\pm$ 4.06} & 68.37 \tiny{$\pm$ 6.51} & 75.36 \tiny{$\pm$ 4.86} & 75.19 \tiny{$\pm$ 2.56} & 80.85 \tiny{$\pm$ 1.23} \\
 & 16 &58.69 \tiny{$\pm$ 6.08} & 70.21 \tiny{$\pm$ 1.94} & 70.29 \tiny{$\pm$ 7.40} & 76.20 \tiny{$\pm$ 2.90} & 75.26 \tiny{$\pm$ 1.07} & 81.25 \tiny{$\pm$ 1.41} \\
                    \midrule
\multirow{4}{*}{Electronics} & 4 &58.77 \tiny{$\pm$ 6.10} & 61.63 \tiny{$\pm$ 7.30} & 64.13 \tiny{$\pm$ 10.34} & 72.82 \tiny{$\pm$ 6.34} & 65.68 \tiny{$\pm$ 6.80} & 74.88 \tiny{$\pm$ 16.59} \\
 & 8 &59.00 \tiny{$\pm$ 5.78} & 66.29 \tiny{$\pm$ 5.36} & 64.21 \tiny{$\pm$ 10.49} & 75.07 \tiny{$\pm$ 3.40} & 68.54 \tiny{$\pm$ 5.61} & 81.29 \tiny{$\pm$ 1.65} \\
 & 16 &67.32 \tiny{$\pm$ 4.18} & 69.61 \tiny{$\pm$ 3.54} & 71.12 \tiny{$\pm$ 7.29} & 75.40 \tiny{$\pm$ 2.43} & 67.84 \tiny{$\pm$ 7.23} & 81.86 \tiny{$\pm$ 1.56} \\
                    \midrule
\multirow{3}{*}{Kitchen} & 4 &56.93 \tiny{$\pm$ 7.10} & 63.07 \tiny{$\pm$ 7.80} & 60.53 \tiny{$\pm$ 9.25} & 75.40 \tiny{$\pm$ 6.27} & 62.71 \tiny{$\pm$ 9.53} & 78.35 \tiny{$\pm$ 18.36} \\
 & 8 &57.13 \tiny{$\pm$ 6.60} & 68.38 \tiny{$\pm$ 4.47} & 69.66 \tiny{$\pm$ 8.05} & 75.13 \tiny{$\pm$ 7.22} & 70.19 \tiny{$\pm$ 6.42} & 84.88 \tiny{$\pm$ 1.12} \\
 & 16 &68.88 \tiny{$\pm$ 3.39} & 75.17 \tiny{$\pm$ 4.57} & 77.37 \tiny{$\pm$ 6.74} & 80.88 \tiny{$\pm$ 1.60} & 71.83 \tiny{$\pm$ 5.94} & 85.27 \tiny{$\pm$ 1.31} \\
                    \bottomrule
\end{tabular}
\caption{Domain transfer evaluation (accuracy) on Sentiment classification datasets.}
\label{tab:amazon}
\end{table*}
\begin{table*}[ht]
\centering
\begin{tabular}{cccccccccc}
\Xhline{2\arrayrulewidth}
 & MNLI\small{(m/mm)} & QQP & QNLI & SST-2 & CoLA & MRPC & RTE & SNLI & Average\\ \Xhline{2\arrayrulewidth}
MT-BERT &  82.11 & 89.92 & 89.62 & 90.7 & 81.30 & 84.56 & 78.34 & 89.97 & \textbf{85.82} \\ \hline
\end{tabular}
\caption{Dev-set accuracy on the set of train tasks for multi-task BERT.}
\label{tab:multidev}
\end{table*}
\noindent Table \ref{tab:amazon} shows the accuracy on all the four amazon sentiment classification tasks.

\noindent Table \ref{tab:multidev} shows the dev-set accuracy of our trained MT-BERT model on the various training tasks.

\begin{figure*}[htb!]
    \centering
    \includegraphics[width=\textwidth]{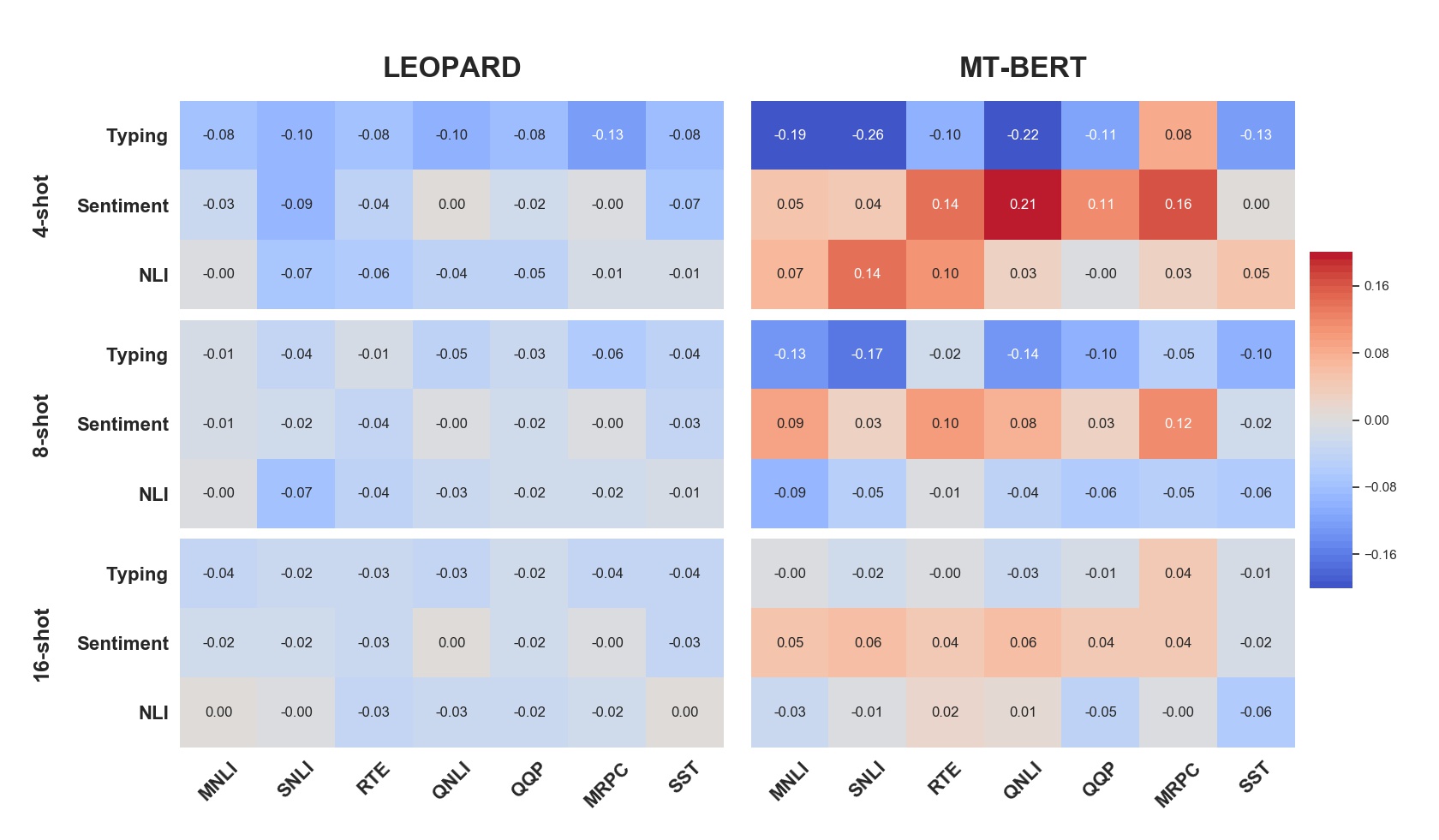}
    \caption{Analyzing target task performance as a function of training tasks (best viewed in color). Heatmaps on the left are for \method and on the right are for MT-BERT. 
  Each column represents one held-out training task (name on $x$-axis) and each row corresponds to one target task (name on $y$-axis). 
  Each cell is the relative change in performance on the target task when the corresponding training task is held-out, compared to training on all the train tasks.
  Dark blue indicates large drop, dark red indicates large increase and grey indicates close to no change in performance.
  In general, \method's performance is more consistent compared to MT-BERT indicating that meta-training learns more generalized initial parameters compared to multi-task training.}
    \label{fig:heldout_full}
\end{figure*}

\noindent Figure~\ref{fig:heldout_full} shows the target task performance as a function of training tasks for all $k$. Note that the effect of training tasks starts to decrease as $k$ increases.

\section{Hyperparameters}
\label{sec:hyper}
Table \ref{tab:params_our} shows the hyper-parameter search range as well as the best hyper-parameters for MT-BERT, Proto-BERT and \method. We use same hyperparameters for prototypical networks except those not relevant to them. 
For fine-tuning we separately tune number of iterations, 
and batch size for each k shot for all the baselines. 
We also tuned warm-up \cite{devlin2018bert} in $\{0, 0.1\}$ and used 0.1 for all the methods.
For MT-BERT we found 10 epochs , batch size 8 to be best for 4-shot, 5 epochs, batch size 8 to be best for 8-shot and 5 epoch with 16 batch size gave the best performance for 16 shot. For $\mbox{MT-BERT}_{\mbox{softmax}}$ we found 125 epoch, batch size 4 to be best for 4-shot, 125 epochs, batch size 4 to be best for 8-shot and 125 epochs with batch size 4 gave the best performance for 16-shot. For $\mbox{BERT}_{\mbox{base}}$ 10 epochs, batch size 8 for 4 shot, 5 epochs, 16 batch size for 8 shot and 10 epochs, batch size 16 for 16 shot gave the best performance. For $\mbox{MT-BERT}_{\mbox{reuse}}$ we found 10 epochs , batch size 8 to be best for 4-shot, 5 epochs, batch size 8 to be best for 8-shot and 5 epoch with 16 batch size gave the best performance for 16 shot.
Note, for \method we use learned per-layer learning rates with SGD. We use the following values: 150 epochs for 4-shot, 100 epochs for 8-shot, 100 epochs for 16-shot. 
 

\begin{table}[htb!]
\centering
\resizebox{\linewidth}{!}{%
\begin{tabular}{ccccc}
\Xhline{2\arrayrulewidth}
Parameter & Search Space & MT-BERT & Proto-BERT & \method \\ \Xhline{2\arrayrulewidth}
Attention dropout & [0.1, 0.2, 0.3] & 0.2 & 0.3 & 0.1 \\ \hline
Batch Size & [16, 32] & 32 & 16 & 10 \\ \hline
Class Embedding Size & [128, 256] & --- & 256 & 256  \\ \hline
Hidden Layer Dropout & [0.1, 0.2, 0.3] & 0.1 & 0.2 & 0.1\\ \hline
Inner Loop Learning Rate & --- & --- & --- & Meta-SGD (per-layer)\\ \hline
Min Adapted Layer ($\nu$) & [0, 5, 8, 10, 11] & --- & --- & 0\\ \hline
Outer Loop Learning Rate & [1e-4, 1e-5, 2e-5, 4e-5, 5e-5] &  2e-05 & 2e-05 & 1e-05\\ \hline
Adaptation Steps ($G$) & [1, 4, 7] & --- & --- & 7\\ \hline
Top layer [CLS] dropout & [0.45, 0.4, 0.3, 0.2, 0.1] & 0.1 & 0.2 & 0.1 \\ \hline
Train Word Embeddings(Inner Loop) & [True, False] & --- & --- & True\\ \hline
Data Sampling & [Square Root, Uniform] &Square Root & Square Root & Square Root \\ \hline
Lowercase text & False & False & False & False \\ \hline
\end{tabular}
}
\caption{Hyper-parameter search space and best hyper-parameters for all models.}
\label{tab:params_our}
\end{table}

\end{appendix}

\end{document}